\documentclass{bmvc2k}
\usepackage{amssymb}
\usepackage{multirow} 
\usepackage{amsmath}  

\title{Aristotelian Manifolds: Leveraging Platonic Perceptual Features for Backpropagation Free Rapid Concept Learning}

\addauthor{Michael Karnes}{karnes.30@osu.edu}{1}
\addauthor{Alper Yilmaz}{yilmaz.15@osu.edu}{1}

\addinstitution{
 Photogrammetric Computer Vision Lab\\
 The Ohio State University\\
 Columbus, OH, USA
}

\runninghead{AUTHOR(S)}{M. Karnes, A. Yilmaz}

\begin{document}

\maketitle

\begin{abstract}
This paper formalizes and systematically characterizes Aristotelian Manifolds, a generalized structural framework built upon the Platonic Representation Hypothesis. We position high-capacity foundation models as universal perceptual filters and conduct a comprehensive layer-wise investigation to map how knowledge is functionally synthesized within these latent subspaces. Across diverse architectural paradigms and multi-domain datasets, we rigorously chart the interplay between network depth, dimensionality reduction, and distance metrics. Our characterization reveals that semantic maturation does not follow a singular, monotonic path; instead, different data domains exhibit highly distinct geometric response profiles, characterized by intermediate mound-like peaks for specialized clinical modalities and sigmoidal plateaus for natural visual tasks. By profiling the exact coordinates where these manifolds achieve peak representational efficiency, we establish a predictable taxonomy for layer selection and feature compression. Ultimately, this systematic characterization demonstrates that mapping the internal geometry of frozen representations provides a robust, backpropagation-free, and interpretable framework for understanding and exploiting foundation model latent spaces.
\end{abstract}

\section{Introduction}
\label{sec:intro}
Deep learning has undergone a fundamental paradigm shift toward universal representations, fueled by the introduction of the Platonic Representation Hypothesis (PRH) \cite{huh2024platonic}. This framework suggests that diverse neural architectures, when trained on massive and varied datasets, naturally converge toward a shared geometric manifold. Under this view, the pretrained latent spaces of foundation models serve as a universal perceptual filter, capable of organizing complex visual data into stable, structural "Platonic Forms" without requiring task-specific backpropagation. However, while the PRH establishes the existence of this global convergence, it lacks a systematic methodology for leveraging these properties to efficiently synthesize such forms into practical, resource-constrained classification applications.

To bridge this gap, we introduce the concept of Aristotelian Manifolds, structured subspaces generated from the representation signals of the Platonic perceptual filter organized into explicit Concept Dictionaries. By replacing opaque, gradient-based decision layers with distance-based logic via $k$-Nearest Neighbors ($k\text{NN}$), this framework transforms black-box inferences into interpretable geometric decisions. This synthesis enables rapid object modeling that reflects the Aristotelian view of human knowledge: the active, structured organization of sensory experience through innate cognitive primitives.

Crucially, our work maps the heterogeneous response profiles of diverse datasets across the individual layers of prominent modern architectures. Through a systematic layer-wise investigation, we demonstrate that the concept-ready state is not a fixed destination restricted to the final layer. Instead, we observe distinct architectural trends, such as mound-like peaks in intermediate-to-late layers for specialized medical data and sigmoidal plateaus for general object recognition. By identifying the layer where an Aristotelian Manifold reaches peak efficiency, we enable optimal layer selection for strategic network trimming and significant dimensionality reduction, offering a clear pathway for computational optimization and predictable deployment in resource-constrained environments.

The primary contribution of this work is a systematic evaluation of Aristotelian Manifold performance across four interconnected dimensions. Specifically, we conduct an extensive comparative analysis across diverse architectural paradigms, spanning VGG, ResNet, ViT, and DINOv2. Within these architectures, we map the layer-wise dynamics of representation quality across individual hidden layers, characterizing the layer-to-concept response. Furthermore, we investigate the impact of aggressive dimensionality reduction on downstream classification accuracy, alongside a comprehensive exploration of various distance metrics to determine the optimal geometric space for distance-based inference.

\section{Related Work}

\subsection{The Platonic Filter and Perceptual Forms}
The Platonic Representation Hypothesis (PRH) posits that the latent representations of deep visual and language models converge toward a shared geometry, regardless of differences in architecture or training objectives \cite{huh2024platonic}. This empirical phenomenon was recently given a rigorous mathematical foundation by \cite{ziyin2025proo}, who proved that under stochastic gradient descent (SGD), networks of varying widths and depths naturally converge to an aligned, universal geometry. Driven by implicit entropic forces inherent to the training dynamics, this mathematical guarantee shows that even disparate hidden layers learn identical representations up to a global rotation. This theoretical framework is further supported by a growing body of empirical literature on cross-latent space stitching, which demonstrates the practical feasibility of directly mapping and aligning these representation spaces \cite{moschella2023relative, bansal2021revisiting, jha2025harnessing}.

This perspective of generalized features builds upon a long lineage of transfer learning. Shortly after the debut of AlexNet, Deep Convolutional Activation Features (DeCAF) were introduced to demonstrate how pretrained deep representations could be successfully leveraged for novel downstream tasks \cite{Donahue2014DeCAF, astounding_decaf, transfer_decaf}. From these foundations, utilizing frozen pretrained models has become standard practice, whether through selective layer fine-tuning, student-teacher distillation, or direct off-the-shelf feature extraction \cite{Tian2020, distill2021, Lora2023, transfersurvey2025}. Notably, \cite{Kornblith2019} formalized this paradigm by benchmarking 16 modern architectures across 12 image classification datasets. By evaluating models as fixed feature extractors (via logistic regression), applying full-model fine-tuning, and contrasting both against training from scratch, they revealed a powerful correlation between a model's upstream performance and its downstream transfer efficacy. This strategy remains central to SOTA architectures; for instance, \cite{tang2025} recently achieved top-tier few-shot learning performance by pairing transfer learning with advanced student-teacher knowledge distillation.

Given the inherent stochasticity of gradient descent, the massive scale of over-parameterized networks, and the vast diversity of training objectives, this convergence toward a universal feature geometry is profoundly remarkable. The implications of the Platonic Representation Hypothesis push the boundaries of standard transfer learning, forcing a deeper inquiry into just how far these universal features can be extended, particularly when navigating the decisions of varying architectures, dimensionality reductions, and distance metrics.

\subsection{Theoretical Framework of Aristotelian Manifolds}
The foundational premise of our work is that the Platonic Representation Hypothesis (PRH) effectively describes a universal perceptual filter \cite{huh2024platonic}. While the PRH establishes that large-scale models naturally converge toward a shared, objective geometry of reality, we argue that this convergence yields the essential, innate cognitive primitives required for structured semantic organization. Rather than treating this universal geometry as a passive final endpoint, we position it as the necessary raw material from which explicit downstream knowledge structures can be systematically organized.

In classical philosophy, Plato posited that concepts ("Forms") exist as an objective, finite reality; thus, learning is not a process of discovering new concepts, but rather of uncovering or recalling these pre-existing truths \cite{Plato1997Republic, Plato1997Phaedo}. Conversely, Aristotle argued that concepts are stabilized through sensory experience, facilitated by an innate capacity for perception. In the Aristotelian view, knowledge is constructed through a bottom-up process of extracting and realizing a universal substance from the accumulation of specific, concrete particulars \cite{AristotleMetaphysics}.

Mirroring this classical shift from abstract idealism to empirical organization, we introduce Aristotelian Manifolds as the structured, task-specific organization of universal Platonic features. An Aristotelian Manifold is delineated not by a static, universal ideal, but by the functional realization of these latent representations when applied to a specific domain. Within this framework, downstream knowledge emerges through the explicit clustering and geometric organization of concrete examples into structured, categorical concepts.

\subsection{Layer-wise Dynamics}
While many works default to utilizing pretrained features from the final layers, largely due to the computational overhead of layer-wise exploration and the assumption that deeper layers yield the richest representations, studies investigating intermediate layers reveal a more nuanced landscape. For instance, an early exploration of DeCAF fine-tuning found that feature transferability varies significantly across a network: early layers remain highly transferable, middle layers are relatively fragile, and deeper layers become strictly task-specific \cite{transfer_decaf}. Conversely, subsequent work employing linear probes to evaluate class separability observed a contrasting trend, demonstrating that feature separability increases consistently with layer depth \cite{alain2017understanding}.

Recent literature indicates that intermediate layers contain the most distinctive representation information, suggesting that final layers become too specialized and lose generalizability for downstream transfer tasks \cite{skean2025layer, batra2026, karnes2026aristotel, karnes2026rethinking}. For instance,  \cite{karnes2026aristotel} introduced the concept of "Aristotelian Representations" by exploring dataset responses across DINOv2 layers using the diverse MedMNIST collection. By pairing high dimensionality reduction with simple k-nearest neighbor (kNN) classification, their approach achieved competitive performance against standard benchmarks. In a closely related study, \cite{karnes2026rethinking} extended this methodology to few-shot learning tasks, demonstrating that layer-aware frozen feature extraction coupled with significant dimensionality reduction can yield state-of-the-art (SOTA) performance.

While these recent studies highlight the latent capabilities of modern pretrained architectures, they stop short of systematically evaluating how these characteristics vary across different model architectures, dimensionality reduction techniques, and distance metrics for visual classification. Our work directly addresses this gap. By exploring the interplay between these three dimensions, we provide a comprehensive evaluation of layer-wise representation dynamics.

\section{Methodology}
\section{Hierarchical Encoding Framework}
The proposed feature modeling framework utilizes a multi-staged pipeline designed to map raw visual inputs into a robust, low-dimensional descriptor space following the methodology of \cite{karnes2026aristotel}. Structurally, the workflow is strictly divided into an Unsupervised Encoding Pipeline and a subsequent Supervised Concept Synthesis Phase. The encoding phase processes input data through three distinct, cumulative representation stages: the Raw Stage, the Component Stage, and the Full Stage.

\subsection{Unsupervised Encoding Pipeline}
The encoding pipeline operates entirely without class labels, leveraging task-agnostic geometric regularities embedded within a frozen foundation model backbone to construct a universal representation language.

\textbf{Raw Stage: Latent Feature Extraction:}
The process begins by distilling global visual primitives from an input image utilizing a frozen backbone. For an input image $x$, global context is extracted from a targeted layer $l$. The high-dimensional global latent vector $z \in \mathbb{R}^{Z}$ is derived by performing global average pooling over the $N$ spatial patch tokens:

\begin{equation}
z = \frac{1}{N}\sum_{i=1}^{N} p_{i,l}
\end{equation}

where $p_{i,l}$ represents the $i$-th patch token at layer $l$. This raw embedding preserves the uncompressed, objective geometry of the converged latent manifold.

\textbf{Component Stage: Linear Manifold Refinement:}
To optimize downstream computational tractability and mitigate sensitivity to low-variance idiosyncratic noise, the raw latent space undergoes unsupervised linear refinement. We compute a Principal Component Analysis (PCA) transform using centered, unlabeled latent vectors from the background dataset. The centered raw feature vector is projected onto the axes of maximum global variance to form the Alphabet vector $a \in \mathbb{R}^{A}$:

\begin{equation}
a = W_{\text{PCA}}^{T}(z - \mu)
\end{equation}

where $W_{\text{PCA}} \in \mathbb{R}^{Z \times A}$ denotes the projection matrix containing the top $A$ principal components and $\mu$ is the empirical mean of the latent distribution. This step filters out non-essential dimensions while preserving key geometric relationships.

\textbf{Full Stage: Unsupervised Vocabulary Quantization:}
The final encoding step maps continuous structural features into a symbolic, distance-based representation language. Using $K$-means clustering executed on the unlabeled Alphabet vectors, we establish a fixed Vocabulary consisting of $V$ prototypical centroids, $\mathcal{V} = \{v_1, v_2, \dots, v_V\}$. A localized Word encoding vector $d \in \mathbb{R}^V$ is constructed by calculating the Euclidean distance from the refined Alphabet vector $a$ to each cluster centroid:

\begin{equation}
d_j = \|a - v_j\|_2, \quad \forall j \in \{1, \dots, V\}
\end{equation}

To systematically investigate the structural interaction between continuous and discrete features, the final composition of the Full Encoding vector $s$ is explicitly controlled as Non-Concatenated and Concatenated. For Non-Concatenated, the description vector is the quantized Word vector $d$ as an independent representation space:
\begin{equation}
s = d
\end{equation}
For the Concatenated case, the description vector is generated via the direct concatenation ($\oplus$) of the continuous Alphabet vector $a$ and the distance-based Word vector $d$, a hybrid descriptor ($s \in \mathbb{R}^{A+V}$):
\begin{equation}
s = [a \oplus d]
\end{equation}

\subsection{Supervised Concept Synthesis Phase}
To maximize class separability within the localized representation space, the encoding vectors are mapped into a supervised discriminant space. Labeled training samples for each class $c$ are aggregated into class-specific collections $\mathcal{S}_{c} = \{s_{1,c}, s_{2,c}, \dots, s_{n,c}\}$. The Linear Discriminant Analysis (LDA) projection matrix $W_{\text{LDA}}$ is derived by solving the generalized eigenvalue problem $S_B w = \lambda S_W w$. The final aligned concept vector $\tilde{s} \in \mathbb{R}^m$ is calculated as:

\begin{equation}
\tilde{s} = W_{\text{LDA}}^{T}s
\end{equation}

where the reduced target dimensionality $m$ is strictly bounded by the number of target categories such that $m \le C - 1$ for a classification task with $C$ classes.

These supervised projections populate the final Concept Dictionary. This dictionary is used to perform k-NN classification for inference with either Mahalanobis distance, Euclidean distance, or Cosine similarity.

\section{Experimental Setup}
\subsection{Datasets}
We evaluate our framework across two benchmark ecosystems to test cross-domain generalization and few-shot efficiency under extreme constraints: MedMNIST v2 \cite{yang2023medmnist} and standard few-shot learning (FSL) visual classification datasets. For high-stakes medical diagnosis, we utilize eleven 2D datasets from the MedMNIST v2, omitting the multi-label ChestMNIST for single-label consistency. For open-domain FSL benchmarks, we evaluate on four vision datasets: miniImageNet \cite{SunLCS2019MTL,LiuMTLDownload,Russakovsky2015}, tieredImageNet \cite{SunLCS2019MTL,LiuMTLDownload,Russakovsky2015}, CIFAR-FS \cite{zenodo_cifar100,Krizhevsky09}, and FC100 \cite{SunLCS2019MTL,LiuMTLDownload,Krizhevsky09}. Together, these benchmarks bridge natural imagery and fine-grained pathology across diverse sample sizes, resolutions, and granularities to demonstrate the framework's broad utility.

\subsection{Architecture and Layer-wise Characterization Sweep Protocol}
To systematically analyze feature maturation and identify optimal representational depths, we execute a layer-wise evaluation for all three encoding stages across VGG \cite{simonyan2015very}, ResNet \cite{he2016deep}, ViT \cite{dosovitskiy2021an}, and DINOv2 \cite{oquab2024dinov} architectures. Downstream evaluation configurations and the sequential hyperparameter routing stages are structured in Table~\ref{tab:experimental_protocol}.

\begin{table}[ht]
\centering
\caption{Experimental Protocol and Hyperparameter Sweep Stages}
\label{tab:experimental_protocol}
\footnotesize 
\setlength{\tabcolsep}{4pt} 
\begin{tabular}{llp{5.8cm}} 
\hline
\textbf{Category} & \textbf{Parameter / Stage} & \textbf{Specification / Search Grid} \\ \hline
\multirow{4}{*}{\textbf{Evaluation Setup}} 
& Unsupervised Training & 1000 random images \\
& Supervised Training   & 64 images/class \\
& Test Evaluation       & 200 random images \\
& Inference Classifier  & $k\text{NN}$ ($k=3$), Mahalanobis distance \\ \hline
\multirow{3}{*}{\textbf{Sweep Stages}} 
& Stage 1: Raw Encoding & Uncompressed sweep across all backbone layers. \\
& Stage 2: Component Encoding & Layer sweep with PCA: $A \in \{64, 128, 256, 512\}$. \\
& Stage 3: Full Encoding & Layer sweep; PCA $A = 512$; $K$-means $V \in \{64, 128, 256, 512\}$, Non-Concatenated. \\ \hline
\end{tabular}
\end{table}

\subsection{Comparison Evaluation Protocols}
To rigorously assess the generalizability and adaptation efficiency of the hierarchical representation pipeline, we evaluate the system under two distinct testing paradigms tailored to their respective domains: an All-Way Evaluation Protocol for the domain-specific MedMNIST v2, and a standard Episodic Few-Shot Learning Protocol for the open-domain FSL vision benchmarks. The results of Architecture and Layer-wise Characterization Sweep Protocol were used to select the top architecture looking at the top mean accuracy across datasets across the group and layers and hyperparameters for each individual dataset case within the groups. A comparative overview of these structural designs is summarized in Table~\ref{tab:fsl_protocols}.

\begin{table}[ht]
\centering
\caption{Structural Comparison Between Evaluation Protocols}
\label{tab:fsl_protocols}
\footnotesize 
\setlength{\tabcolsep}{4pt} 
\begin{tabular}{lp{4.2cm}p{4.2cm}}
\hline
\textbf{Metric} & \textbf{MedMNIST Paradigm } & \textbf{FSL Benchmark Paradigm} \\ \hline
Backbone Model        & DINOv2 ViT-Large/14 & ViT-Large/16 \\
Evaluation Style      & All-Way (All classes jointly) & Episodic $N$-Way (5-Way) \\
Data Regime           & 512-Shot Fixed Target & 1-Shot and 5-Shot Regimes \\
Statistical Depth     & 5 random support trials & 600 episodes ($\text{Seed} = 42$) \\
Query Pool            & Full Test Split ($n_{\text{eval}} = -1$) & 15 samples/class/episode \\
Unsupervised Pool     & $\le$ 5,000 samples/class & 500 unlabeled background images \\
Data Augmentation     & None & Dihedral ($D_4$) (7 copies/sample) \\
Classifier            & LDA Aligned + $k\text{NN}$ ($k=15$) & $k\text{NN}$ ($k=15$) Direct Inference \\
Stage 3 Vector        & Concat-Enabled ($s = [a \oplus d]$) & Concat-Enabled ($s = [a \oplus d]$) \\ \hline
\end{tabular}
\end{table}

\subsection{MedMNIST All-Way Evaluation Protocol}
The MedMNIST v2 datasets are evaluated using a macro-level, all-way classification architecture that strictly builds and verifies the modeling constructs across a single, unified label matrix rather than breaking tasks down into localized meta-learning sub-episodes, matching \cite{karnes2026aristotel}. The execution pipeline operates in two distinct execution blocks:

\textbf{Unsupervised Encoding Phase:} Unsupervised compression maps are established using the optimal layer and parameters for each stage and an unlabeled training allocation of up to $5,000$ images per class.

\textbf{Supervised Concept Synthesis Phase:} The system evaluates model adaptability with a 512-shot threshold per class. To minimize statistical bias, the model executes across $5$ independent trials featuring randomized redraws of the labeled support exemplars. During inference, a downstream LDA projection is added on top of the representations before feeding vectors into the $k\text{NN}$ module ($k=15$). Testing metrics are calculated across the Mahalanobis, Euclidean, and Cosine distance landscapes mapped directly across the full test collection of the dataset.

\subsection{Standard Episodic FSL Benchmark Protocol}
For standard visual recognition benchmarks (miniImageNet, tieredImageNet, CIFAR-FS, FC100), the model is benchmarked against meta-learning variants through a traditional episodic framework. Each model framework is evaluated across the Mahalanobis, Euclidean, and Cosine distance landscapes.

The few-shot learning framework is evaluated using an episodic, $N$-way classification architecture that strictly constructs and verifies the modeling across localized task spaces rather than a single, unified label matrix, matching \cite{karnes2026rethinking}. The execution pipeline operates in two distinct, sequential execution blocks:

\textbf{Unsupervised Encoding Phase:} The unsupervised compression maps are established entirely without class labels prior to episodic evaluation. This upfront phase utilizes an allocation pool of $500$ task-agnostic reference frames, stratified across the base training classes, to derive global PCA transformations and initialize $K$-Means centroid dictionaries. These training classes remain strictly disjoint from the downstream evaluation targets. Crucially, this encoder-fitting step is executed once before meta-testing begins; the resulting embedding space is frozen and reused across all subsequent tasks, incurring a zero marginal computational cost per episode.

\textbf{Supervised Concept Synthesis Phase:} To evaluate model adaptability in low-data regimes, the system samples $600$ randomized episodes anchored to a fixed initialization parameter ($\text{seed} = 42$). Each individual episode draws an $N$-way support set from the test collection under both 1-shot and 5-shot regimes. To guard against sample variance, these raw support exemplars are geometrically expanded using Dihedral Group ($D_4$) data transformations. Each source image generates $7$ unique spatial variants constructed via independent vertical, horizontal, and $90^\circ$ rotational flips under uniform sampling probabilities ($p=0.5$), completely free of additive noise. 

This geometric expansion constructs a strictly bounded, dense episodic concept dictionary comprising exactly $40$ concept vectors for the 1-shot setting ($5\text{ classes} \times 1\text{ shot} \times 8\text{ total variants}$) and $200$ concept vectors for the 5-shot setting ($5\text{ classes} \times 5\text{ shots} \times 8\text{ total variants}$). Unlike MedMNIST All-Way, no LDA projection is used due to the extreme constraint on the concept dictionary size.

\textbf{Episodic Evaluation Inference:} Each individual episode uses $15$ unlabeled evaluation queries per target class ($75$ queries total per episode), drawn via a strict no-replacement strategy from the sampled test classes. These query vectors are fed directly into a $K$-Nearest Neighbors module ($K=15$). Testing metrics are derived across the Mahalanobis, Euclidean, and Cosine distance metrics and mapped explicitly against the episodic concept dictionary vectors. Task accuracy is computed as the mean prediction success over these $75$ evaluation queries for each respective distance metric.

\section{Experimental Results}
In this section, we present a systematic empirical evaluation of the hierarchical representation framework, analyzing its performance boundaries across diverse vision domains and architectural depths. Our empirical validation is structured into three specialized analytical trajectories:

\textbf{Models Survey:} A macroscopic cross-architectural assessment evaluating the performance of VGG, ResNet, ViT, and DINOv2 backbones across the entire MedMNIST v2 and FSL Benchmarks to isolate optimal foundational embeddings using the Architecture and Layer-wise Characterization Sweep Protocol.

\textbf{Layer-wise Characterization:} A granular depth-wise sweep mapping feature maturation across sequential backbone layers to identify the representational boundaries where compression and clustering performance peak digging deeper into the Architecture and Layer-wise Characterization Sweep Protocol results.

\textbf{Few-Shot Comparison:} A targeted comparative benchmark contrasting our frozen  unsupervised compression maps and localized concept synthesis phase against traditional meta-learning baselines under strict 1-shot and 5-shot regimes using the Comparison Evaluation Protocols.

\subsection{Models Survey}
\begin{figure}[htbp]
    \centering
    \includegraphics[width=0.9\linewidth]{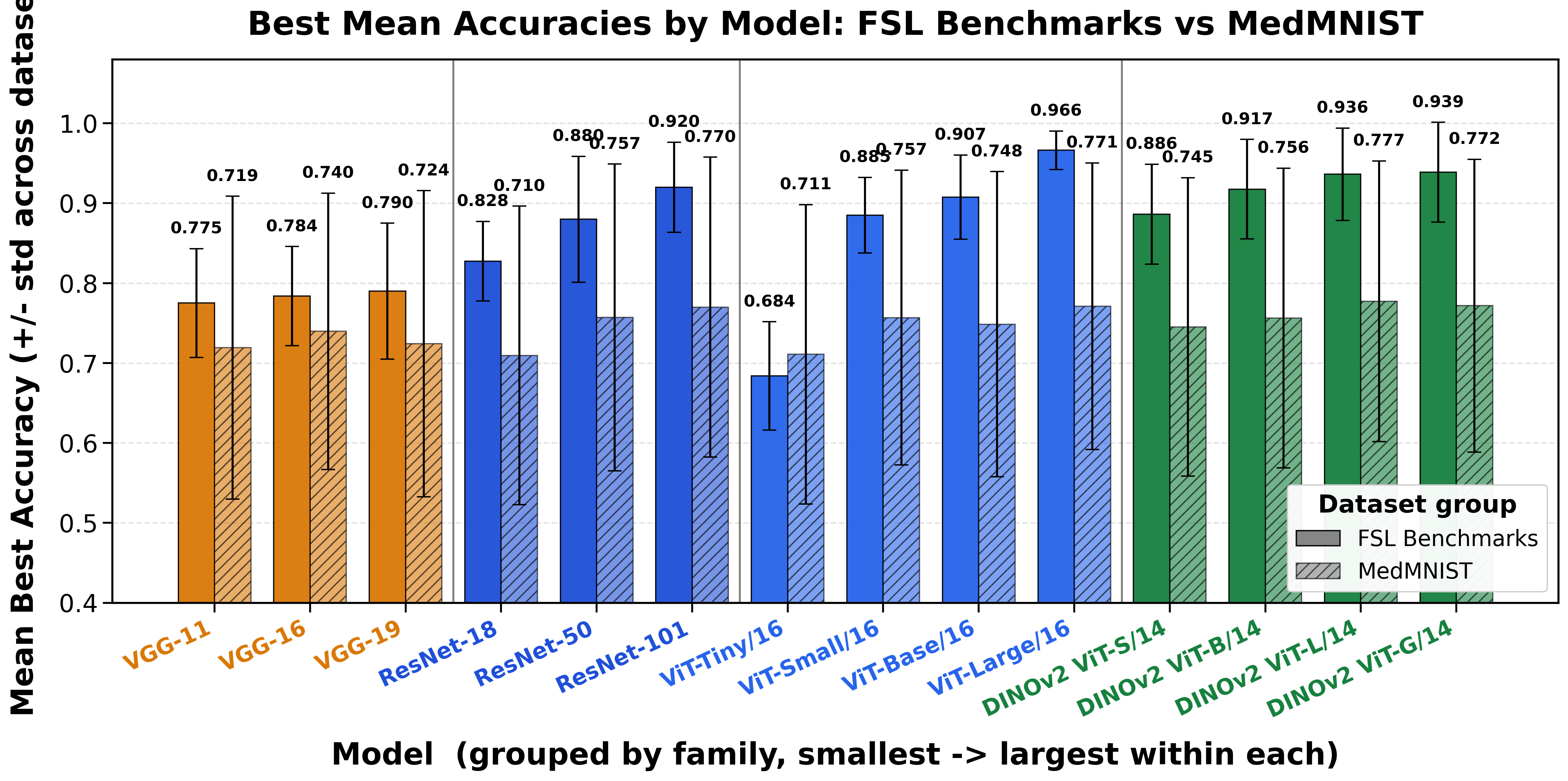}
    \caption{Model performance comparison collapsed across all experimental stages, isolating the single best accuracy achieved by each specific model-dataset pairing across all three stages and layers. Bar heights represent the mean of these peak configurations across the constituent datasets in each group: solid bars denote the FSL Benchmarks group, while hatched bars denote the MedMNIST group. Error bars denote the standard deviation ($\pm$ SD) calculated across the datasets within each model group.}
    \label{fig:model_bars}
\end{figure}

As illustrated in Figure \ref{fig:model_bars}, an evaluation across these architectural families reveals a clear performance bifurcation between open-domain FSL benchmarks and domain-specific MedMNIST v2 tasks. Within the natural image FSL benchmarks, performance scales monotonically with parameter depth and model complexity. This trend culminates in the vision transformer architectures; specifically, ViT-L/16 achieves a peak mean accuracy of $0.966$.

\begin{figure}[htbp]
    \centering
    \includegraphics[width=0.9\linewidth]{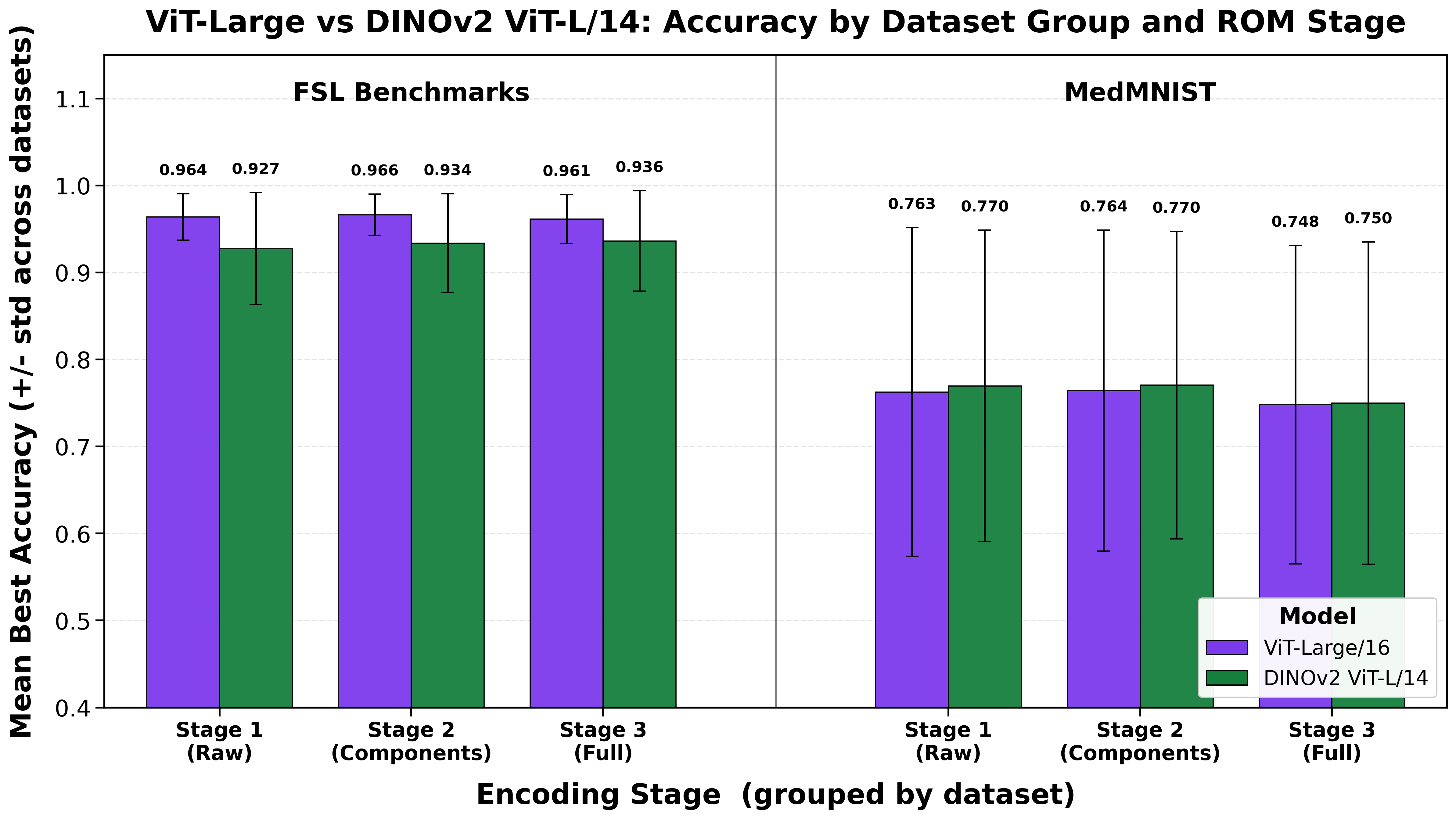}
    \caption{Model performance decomposed by operational stage across benchmark groups. Bar heights represent the mean of the peak layer-wise accuracies computed across all constituent datasets within a given group for each specific (model, group, stage) cell. Error bars denote the standard deviation ($\pm$ SD) of these best-layer configurations across the respective datasets within that specific stage and model cluster.}
    \label{fig:vit_l_stages}
\end{figure}

Figure \ref{fig:vit_l_stages} isolates the performance trajectories of the two leading transformer backbones, ViT-Large/16 and DINOv2 ViT-L/14, decomposed across the three sequential representation encoding stages. For the open-domain FSL benchmarks, the two models exhibit increased average accuracy with compression. The standard ViT-Large backbone peaks shows a notably diminished variance compared to the DINOv2 ViT-L/14 and peaks during Stage 2 component encoding.

When evaluating the same stage transitions on MedMNIST v2 in Figure \ref{fig:vit_l_stages}, the difference in performance across all compression tiers is less significant than the FSL benchmarks. Moving from raw, high-dimensional layer extractions (Stage 1) through PCA dimensionality reduction (Stage 2) to highly quantized $K$-means centroid allocations (Stage 3) yields near equivalent results for both models.  This stability indicates that the structural limitations preventing higher classification accuracy in are not caused by lossy downstream compression for these types of datasets.

\subsection{Layer-wise Characterization}
To uncover the precise evolutionary dynamics of features across deep neural network hierarchies, we conduct a continuous layer-by-layer performance evaluation. This layer-wise trajectory mapping explicitly tracks how different levels of semantic abstraction respond to compression and task-specific clustering pipelines.

\begin{figure}[htbp]
    \centering
    \includegraphics[width=0.9\linewidth]{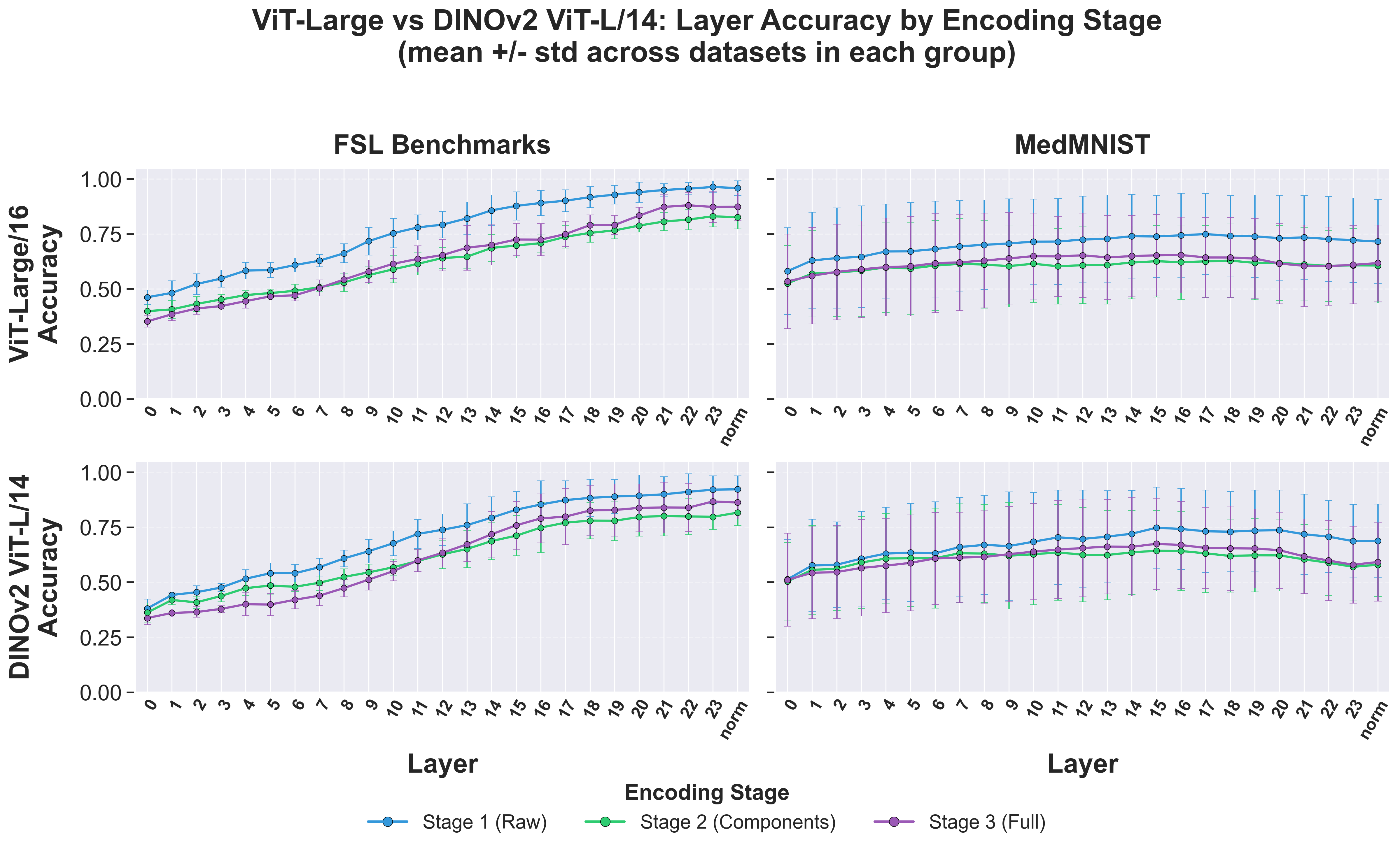}
    \caption{Layer-wise accuracy trends across the complete transformer hierarchy broken down by encoding stages. Each data point represents a hierarchical aggregation: hyperparameter sweeps ($A \in \{64, 128, 256, 512\}$ principal components in Stage 2; $V \in \{64, 128, 256, 512\}$ vocabulary clusters in Stage 3) are condensed into a single mean layer score per dataset, and the final line coordinates plot the group-level mean across all constituent datasets. Vertical error bars represent the standard deviation ($\pm$ SD) of the mean accuracies averaged across the hyperparameters across datasets within that evaluation group.}
    \label{fig:layerwise_plot}
\end{figure}

As detailed in Figure \ref{fig:layerwise_plot}, the continuous layer trajectories expose a stark contrast in feature maturation curves between the two evaluation domains. For the open-domain FSL benchmarks (left column), both the ViT-Large/16 and the DINOv2 ViT-L/14 exhibit a uniform, highly stable upward slope. This continuous ascent demonstrates that for natural imagery, successive transformer layers consistently accumulate generalized, linearly separable semantic features.

Analyzing the interaction between architectural depth and downstream representation compression in Figure \ref{fig:layerwise_plot} reveals an intriguing stratification of the stages. In the natural imagery domain, Stage 1 (Raw) consistently maintains a dominant gap across the entire network hierarchy. As representations are subjected to Stage 2 (Components) and Stage 3 (Full) compression, a systematic performance gap emerges in the middle layers, where the highly quantized Stage 3 trajectory falls increasingly below the raw feature curves. However, as features approach the final exit layers, this compression gap narrows. This indicates that late-stage features possess an inherent geometric robustness, allowing them to undergo PCA reduction and vocabulary clustering with less degradation to downstream episodic classification.

In sharp contrast, the MedMNIST v2 trajectories (right column of Figure \ref{fig:layerwise_plot}) completely reject the monotonic scaling laws observed in open-domain tasks. For the medical pathology datasets, accuracies rise sharply across the early structural blocks before hitting a rigid performance ceiling. For DINOv2 ViT-L/14 specifically, performance actually degrades in the deeper layers, forcing a visible downward slope. This behavior indicates that the deepest layers of models optimized on natural imagery actively discard fine-grained, localized visual primitives in favor of high-level semantic abstractions that are unsuited for clinical anomaly detection.

The right-hand panels of Figure \ref{fig:layerwise_plot} also display exceptionally wide, overlapping vertical error bars across the complete layer spectrum for the MedMNIST v2 group. This massive standard deviation reflects the fundamental structural diversity within the MedMNIST v2, where distinct imaging modalities exhibit entirely independent optimal layer requirements. Specific discussion on optimal layer distributions are included in the Supplemental Materials.

\subsection{Few-Shot Comparison}
To evaluate model adaptability under controlled data constraints, we examine the interaction between our sequential compression maps and the downstream distance landscapes. We begin by isolating the macro-level All-Way 512-shot paradigm on the MedMNIST v2 before contrasting these behaviors with standard episodic few-shot benchmarks.

\subsubsection{MedMNIST}
\begin{figure}[htbp]
    \centering
    \includegraphics[width=0.9\linewidth]{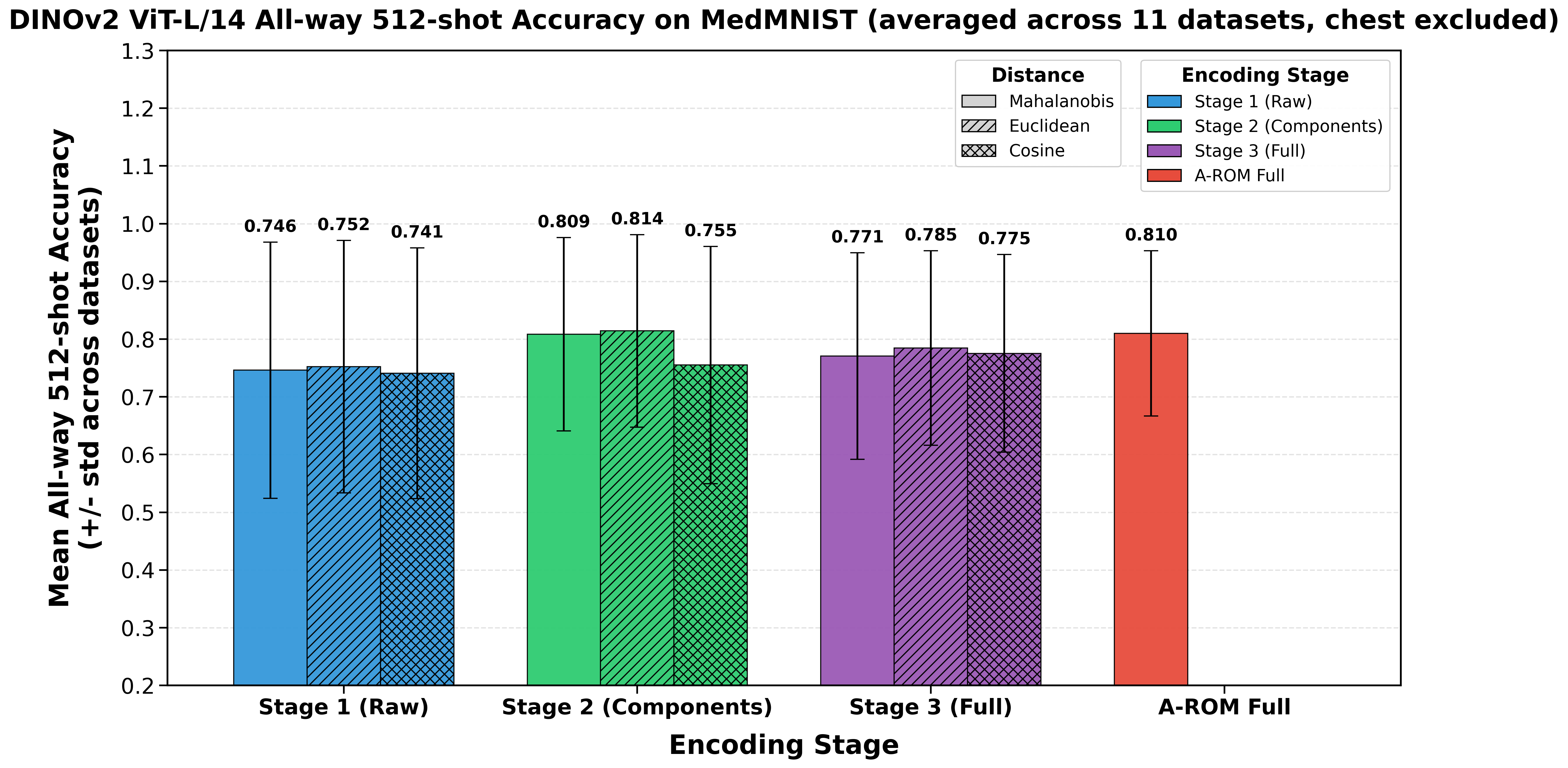}
    \caption{Mean all-way 512-shot classification accuracy across 11 MedMNIST v2 datasets (excluding multi-label ChestMNIST) evaluated using a frozen DINOv2 ViT-L/14 backbone. Performance is broken down across the three encoding stages (Stage 1: Raw, Stage 2: Components , Stage 3: Full) and three distance metrics (Mahalanobis, Euclidean, Cosine). Bar heights show the unweighted mean computed across the 11 individual dataset accuracy scores (where each dataset score represents an underlying 5-trial randomized execution average). Error bars denote the standard deviation ($\pm$ SD) across the dataset cohort, capturing macro-level domain/modality variability rather than intra-trial fluctuations. The A-ROM Full shows the performance with 512 Few-Shot from \cite{karnes2026aristotel}.}
    \label{fig:med_fsl_bars}
\end{figure}

As illustrated by the multi-metric paths in Figure \ref{fig:med_fsl_bars}, the geometric properties of the distance metrics exert a systematic influence on classification accuracy across all three stages. Across Stage 1 (Raw), Stage 2 (Components), and Stage 3 (Full) pipelines, Euclidean distance consistently establishes the dominant performance envelope. This advantage peaks within the Stage 2 principal component space, where Euclidean matching achieves the highest unweighted group mean of 81.4\%, out performing \cite{karnes2026aristotel} 512-Shot mean performance of 81.0\%. Conversely, Cosine distance registers as the lower-bound baseline across two of the three stages. A detailed dataset perspective is shared in the Supplementary Materials.

The structural transitions between the encoding stages in Figure \ref{fig:med_fsl_bars} demonstrate that unsupervised dimensionality reduction acts as an implicit regularizer for out-of-domain embeddings. Moving from high-dimensional uncompressed layers in Stage 1 to the optimized Stage 2 PCA space yields a performance boost across all evaluated metrics, lifting the Mahalanobis, Euclidean, and Cosine landscapes. Strikingly, transitioning further into the highly quantized, discrete vocabulary clusters of Stage 3 preserves this elevated baseline with negligible degradation, maintaining a stable plateau tier across Mahalanobis, Euclidean, and Cosine spaces. This profile demonstrates that unsupervised vocabulary quantization successfully discards redundant, out-of-distribution noise without diluting the core structural features necessary for downstream medical categorization.

Finally, the prominent vertical error bars spanning every evaluation block in Figure \ref{fig:med_fsl_bars} reflect severe cross-dataset volatility across the 11 medical cohorts. The fact that this wide standard deviation remains virtually the same across all three stages and all three distance landscapes indicates that the cross-modality generalization bottleneck is structurally embedded within the frozen foundational features of the backbone itself, completely dominating any downstream metric or compression choices.

\subsubsection{FSL Benchmarks}

To evaluate the competitive boundary of the proposed framework, we analyze the quantitative benchmarks on standard visual recognition datasets as detailed in Table \ref{tab:fsl_imgnet_results_updated} and Table \ref{tab:cifar_fc100_results_updated}. Across MiniImageNet, TieredImageNet, the Mahalanobis-based distance metric configuration generates SOTA performance. The empirical results demonstrate that the Stage 2 Mahalanobis alignment peaks at $89.93\%$ and $97.89\%$ on MiniImageNet and $94.53\%$ and $98.50\%$ on TieredImageNet for 1-shot and 5-shot respectively, setting the leading performance metrics across these foundational evaluations.

\begin{table*}[htb]
\begin{center}
\begin{scriptsize}
\begin{tabular}{|l|l|c|c|c|c|}
\hline
Method & Backbone & \multicolumn{2}{c|}{MiniImageNet} & \multicolumn{2}{c|}{TieredImageNet} \\
\cline{3-6}
 &  & \textbf{5-way 1-shot} & \textbf{5-way 5-shot} & \textbf{5-way 1-shot} & \textbf{5-way 5-shot} \\
\hline\hline
SYNTRANS \cite{tang2025} & ResNet-12 & 76.20 $\pm$ 0.69 & 86.12 $\pm$ 0.54 & 79.69 $\pm$ 0.81 & 87.78 $\pm$ 0.60 \\
SYNTRANS \cite{tang2025} & ViT-S & 81.30 $\pm$ 0.61 & 89.96 $\pm$ 0.42 & 84.31 $\pm$ 0.54 & 91.73 $\pm$ 0.44 \\
\hline
RGEE-Raw \cite{karnes2026rethinking} & DINOv2-L & 83.36 $\pm$ 0.75 & 96.51 $\pm$ 0.24 & 83.61 $\pm$ 0.84 & 94.96 $\pm$ 0.40 \\
RGEE-PCA 512 \cite{karnes2026rethinking} & DINOv2-L & 84.11 $\pm$ 0.74 & 96.18 $\pm$ 0.26 & 84.01 $\pm$ 0.81 & 94.50 $\pm$ 0.42 \\
RGEE-PCA 256 \cite{karnes2026rethinking} & DINOv2-L & 84.16 $\pm$ 0.78 & 95.34 $\pm$ 0.31 & 82.23 $\pm$ 0.81 & 93.09 $\pm$ 0.46 \\
RGEE-PCA 128 \cite{karnes2026rethinking} & DINOv2-L & 82.72 $\pm$ 0.83 & 93.94 $\pm$ 0.40 & 79.89 $\pm$ 0.86 & 91.55 $\pm$ 0.49 \\
\hline
Ours-Mahalanobis-Stage 1 (Raw) & ViT-Large/16 & 89.87 $\pm$ 0.64 & \textbf{\underline{97.89}} $\pm$ 0.16 & 94.25 $\pm$ 0.51 & \textbf{\underline{98.50}} $\pm$ 0.21 \\
Ours-Mahalanobis-Stage 2 (Comp.) & ViT-Large/16 & \textbf{\underline{89.93}} $\pm$ 0.63 & 97.74 $\pm$ 0.17 & \textbf{\underline{94.53}} $\pm$ 0.52 & 97.44 $\pm$ 0.25 \\
Ours-Mahalanobis-Stage 3 (Full) & ViT-Large/16 & 66.37 $\pm$ 0.91 & 92.12 $\pm$ 0.35 & 89.74 $\pm$ 0.69 & 97.02 $\pm$ 0.28 \\
\hline
Ours-Euclidean-Stage 1 (Raw) & ViT-Large/16 & 82.86 $\pm$ 0.87 & 96.99 $\pm$ 0.21 & 89.26 $\pm$ 0.71 & 97.89 $\pm$ 0.25 \\
Ours-Euclidean-Stage 2 (Comp.) & ViT-Large/16 & 80.18 $\pm$ 0.91 & 96.86 $\pm$ 0.22 & 88.23 $\pm$ 0.83 & 97.76 $\pm$ 0.27 \\
Ours-Euclidean-Stage 3 (Full) & ViT-Large/16 & 32.89 $\pm$ 0.71 & 50.62 $\pm$ 0.72 & 64.60 $\pm$ 0.85 & 90.31 $\pm$ 0.50 \\
\hline
Ours-Cosine-Stage 1 (Raw) & ViT-Large/16 & 76.47 $\pm$ 1.15 & 96.99 $\pm$ 0.26 & 83.59 $\pm$ 0.98 & 97.37 $\pm$ 0.27 \\
Ours-Cosine-Stage 2 (Comp.) & ViT-Large/16 & 82.09 $\pm$ 0.84 & 96.82 $\pm$ 0.23 & 90.69 $\pm$ 0.76 & 97.69 $\pm$ 0.26 \\
Ours-Cosine-Stage 3 (Full) & ViT-Large/16 & 35.96 $\pm$ 0.77 & 62.36 $\pm$ 0.75 & 81.83 $\pm$ 0.95 & 96.02 $\pm$ 0.34 \\
\hline
\end{tabular}
\end{scriptsize}
\end{center}
\caption{Comparative results on MiniImageNet and TieredImageNet datasets. Average accuracy (\%) with 95\% confidence intervals.}
\label{tab:fsl_imgnet_results_updated}
\end{table*}

\begin{table*}[htb]
\begin{center}
\begin{scriptsize}
\begin{tabular}{|l|l|c|c|c|c|}
\hline
Method & Backbone & \multicolumn{2}{c|}{CIFAR-FS} & \multicolumn{2}{c|}{FC100} \\
\cline{3-6}
 &  & \textbf{5-way 1-shot} & \textbf{5-way 5-shot} & \textbf{5-way 1-shot} & \textbf{5-way 5-shot} \\
\hline\hline
SYNTRANS \cite{tang2025} & ResNet-12 & 82.58 $\pm$ 0.75 & 89.42 $\pm$ 0.56 & 52.30 $\pm$ 0.75 & 64.91 $\pm$ 0.59 \\
SYNTRANS \cite{tang2025} & ViT-S & 84.64 $\pm$ 0.65 & 90.81 $\pm$ 0.41 & \textbf{56.38} $\pm$ 0.69 & 69.45 $\pm$ 0.54 \\
\hline
RGEE-Raw \cite{karnes2026rethinking} & DINOv2-L & 90.92 $\pm$ 0.58 & \textbf{97.66} $\pm$ 0.19 & 56.01 $\pm$ 0.92 & 77.67 $\pm$ 0.71 \\
RGEE-PCA 512 \cite{karnes2026rethinking} & DINOv2-L & \textbf{91.35} $\pm$ 0.57 & 97.53 $\pm$ 0.20 & 55.34 $\pm$ 0.89 & 76.00 $\pm$ 0.71 \\
RGEE-PCA 256 \cite{karnes2026rethinking} & DINOv2-L & 90.71 $\pm$ 0.61 & 97.14 $\pm$ 0.23 & 53.56 $\pm$ 0.86 & 73.20 $\pm$ 0.72 \\
RGEE-PCA 128 \cite{karnes2026rethinking} & DINOv2-L & 89.59 $\pm$ 0.64 & 96.45 $\pm$ 0.28 & 51.26 $\pm$ 0.85 & 69.68 $\pm$ 0.75 \\
\hline
Ours-Mahalanobis-Stage 1 (Raw) & ViT-Large/16 & 81.24 $\pm$ 0.73 & \underline{96.65} $\pm$ 0.23 & \underline{52.06} $\pm$ 0.85 & \textbf{\underline{79.28}} $\pm$ 0.66 \\
Ours-Mahalanobis-Stage 2 (Comp.) & ViT-Large/16 & \underline{82.09} $\pm$ 0.71 & 96.12 $\pm$ 0.26 & 47.51 $\pm$ 0.78 & 74.06 $\pm$ 0.69 \\
Ours-Mahalanobis-Stage 3 (Full) & ViT-Large/16 & 69.15 $\pm$ 0.84 & 93.41 $\pm$ 0.36 & 37.60 $\pm$ 0.77 & 66.50 $\pm$ 0.70 \\
\hline
Ours-Euclidean-Stage 1 (Raw) & ViT-Large/16 & 70.77 $\pm$ 0.94 & 90.18 $\pm$ 0.47 & 41.20 $\pm$ 0.78 & 65.03 $\pm$ 0.77 \\
Ours-Euclidean-Stage 2 (Comp.) & ViT-Large/16 & 67.88 $\pm$ 0.93 & 89.55 $\pm$ 0.49 & 38.34 $\pm$ 0.69 & 59.74 $\pm$ 0.75 \\
Ours-Euclidean-Stage 3 (Full) & ViT-Large/16 & 30.72 $\pm$ 0.61 & 53.41 $\pm$ 0.69 & 22.54 $\pm$ 0.45 & 30.24 $\pm$ 0.50 \\
\hline
Ours-Cosine-Stage 1 (Raw) & ViT-Large/16 & 67.02 $\pm$ 1.10 & 90.21 $\pm$ 0.48 & 39.97 $\pm$ 0.83 & 62.89 $\pm$ 0.80 \\
Ours-Cosine-Stage 2 (Comp.) & ViT-Large/16 & 69.07 $\pm$ 0.91 & 90.02 $\pm$ 0.47 & 39.68 $\pm$ 0.69 & 59.43 $\pm$ 0.74 \\
Ours-Cosine-Stage 3 (Full) & ViT-Large/16 & 36.51 $\pm$ 0.69 & 65.87 $\pm$ 0.72 & 23.38 $\pm$ 0.44 & 34.62 $\pm$ 0.55 \\
\hline
\end{tabular}
\end{scriptsize}
\end{center}
\caption{Few-shot classification performance on CIFAR-FS and FC100 datasets. Results report average accuracy (\%) with 95\% confidence intervals.}
\label{tab:cifar_fc100_results_updated}
\end{table*}

The comparative metric breakdown across both tables clearly showcases the structural necessity of the Mahalanobis distance metric when classifying dense episodic concept dictionaries. Across all four datasets, the Mahalanobis metric variants (Stage 1 and Stage 2) systematically outperform their corresponding Euclidean and Cosine baselines. This performance disparity is especially pronounced in the challenging 1-shot settings. For instance, on MiniImageNet 1-shot (Table \ref{tab:fsl_imgnet_results_updated}), Mahalanobis Stage 2 achieves SOTA performance, whereas the Euclidean and Cosine equivalents register substantially lower. This persistent advantage demonstrates that capturing the empirical covariance structure of the embedding dimensions is vital for modeling localized class distributions accurately.

While Stage 2 component optimization provides an implicit regularizing effect that balances or enhances accuracy, transitioning into Stage 3 (Full) vector quantization introduces a severe performance bottleneck within the strict few-shot task. As documented across both Table \ref{tab:fsl_imgnet_results_updated} and Table \ref{tab:cifar_fc100_results_updated}, Stage 3 scores plunge dramatically. In the MiniImageNet 1-shot task, Euclidean Stage 3 performance collapses to $32.89\%$, and Cosine Stage 3 lands at $35.96\%$. While the Mahalanobis metric mitigates this drop, it still degrades noticeably at Stage 3.

Finally, on the challenging FC100 dataset (Table \ref{tab:cifar_fc100_results_updated}), performance scales down uniformly across all methods due to the mini-ImageNet-derived superclass structure. Here, our Mahalanobis Stage 1 pipeline establishes a peak 5-shot accuracy at $79.28\%$, but drops to $52.06\%$ in the 1-shot regime, which falls behind the DINOv2-Large-backed RGEE baseline $56.01\%$.

\section{Discussion}

The empirical evaluations across diverse architectures, representational depths, and validation paradigms reveal fundamental properties of frozen foundational latent spaces. By contrasting open-domain FSL benchmarks against the domain-specific MedMNIST v2, these results map the boundaries of the PRH and illuminate the structural mechanics governing Aristotelian Manifolds.

A macroscopic analysis of the architectural scaling profiles (Figure \ref{fig:model_bars}) exposes a critical divergence in how semantic abstractions mature within deep networks. In the natural imagery domain, downstream classification accuracy scales monotonically with backbone parameters and depth. This linear trajectory with architecture validates the PRH's assertion that scaling expansive foundations forces a convergence toward a shared geometry of natural reality. Though, this linear trajectory across architectures diminishes when applied to specialized domains, the continuously mapped  layer-wise trends in Figure (\ref{fig:layerwise_plot}), show that the intermediate latent features are responsive specialized domains like MedMNIST.

A core contribution of this work is demonstrating the geometric resilience of latent manifolds under lossy, unsupervised transformations. The per-dataset stage breakdowns (Figure \ref{fig:vit_l_stages}) reveals that projecting high-dimensional layer extractions (Stage 1) into compact principal component spaces (Stage 2) can actually enhance linear separability in natural imagery and specialized domains by filtering out out-of-distribution noise. More strikingly, transitioning into highly quantized, discrete vocabulary cluster allocations (Stage 3) preserves the downstream performance boundaries with negligible loss. This absolute stability suggests that the intrinsic dimensionality of localized task concepts is far lower than the raw embedding spaces imply.

While downstream compression yields structural stability, the choice of the localized inference classifier uncovers key geometric properties of the underlying feature vectors. In the All-Way MedMNIST evaluations (Figure \ref{fig:med_fsl_bars}), Euclidean distance establishes the dominant performance envelope, while Cosine similarity marks the performance lower bound. Conversely, within the strict episodic constraints of open-domain FSL benchmarks (Table \ref{tab:fsl_imgnet_results_updated} and Table \ref{tab:cifar_fc100_results_updated}), the Mahalanobis distance demonstrates overwhelming superiority over both Euclidean and Cosine baselines.

\section{Conclusion}

In this paper, we formalize and empirically characterize the framework of Aristotelian Manifolds as an efficient, backpropagation-free alternative for downstream task adaptation. Built upon the foundation of the Platonic Representation Hypothesis, we recast the latent space of high-capacity foundation models as universal perceptual filters. Our comprehensive layer-wise characterization sweeps across multiple architectural families expose a fundamental structural bifurcation between natural visual benchmarks and domain-specific clinical modalities. While natural visual benchmarks scale monotonically to deliver new FSL SOTA milestones of 89.93\% and 97.89\% on miniImageNet and 94.53\% and 98.50\% on tieredImageNet, the MedMNIST datasets exhibit localized, modality-dependent layer peaks where essential primitive features are retained to surpass previous few-shot benchmarks with an 81.4\% mean accuracy.

Furthermore, our exploration of sequential representation encoding mappings demonstrated remarkable geometric resilience under aggressive data compression paradigms. Transitioning from high-dimensional uncompressed layer extractions into linearly refined components and quantized vocabulary allocations incurs negligible performance degradation. By showing that low-dimensional embeddings of intermediate network layers can match or exceed the performance of uncompressed representations of the penultimate layers, this work establishes a predictable blueprint for minimizing downstream computational costs. Aristotelian Manifolds provide a dominant, data-efficient, low-cost, and interpretable pathway for deployment across both natural and specialized image spaces.

\clearpage

\bibliography{egbib}

\clearpage

\appendix
\section*{Supplementary Material}
\section{Overview of Supplementary Material}
This supplementary document extends the main manuscript's findings. Specifically, it contains expanded content on the optimal layer distributions, cross-dataset accuracies,  comprehensive few-shot evaluation on MedMNIST datasets, and detailed layer-wise performances.

\section{Architecture}
\begin{figure}[htbp]
    \centering
    \includegraphics[width=0.9\linewidth]{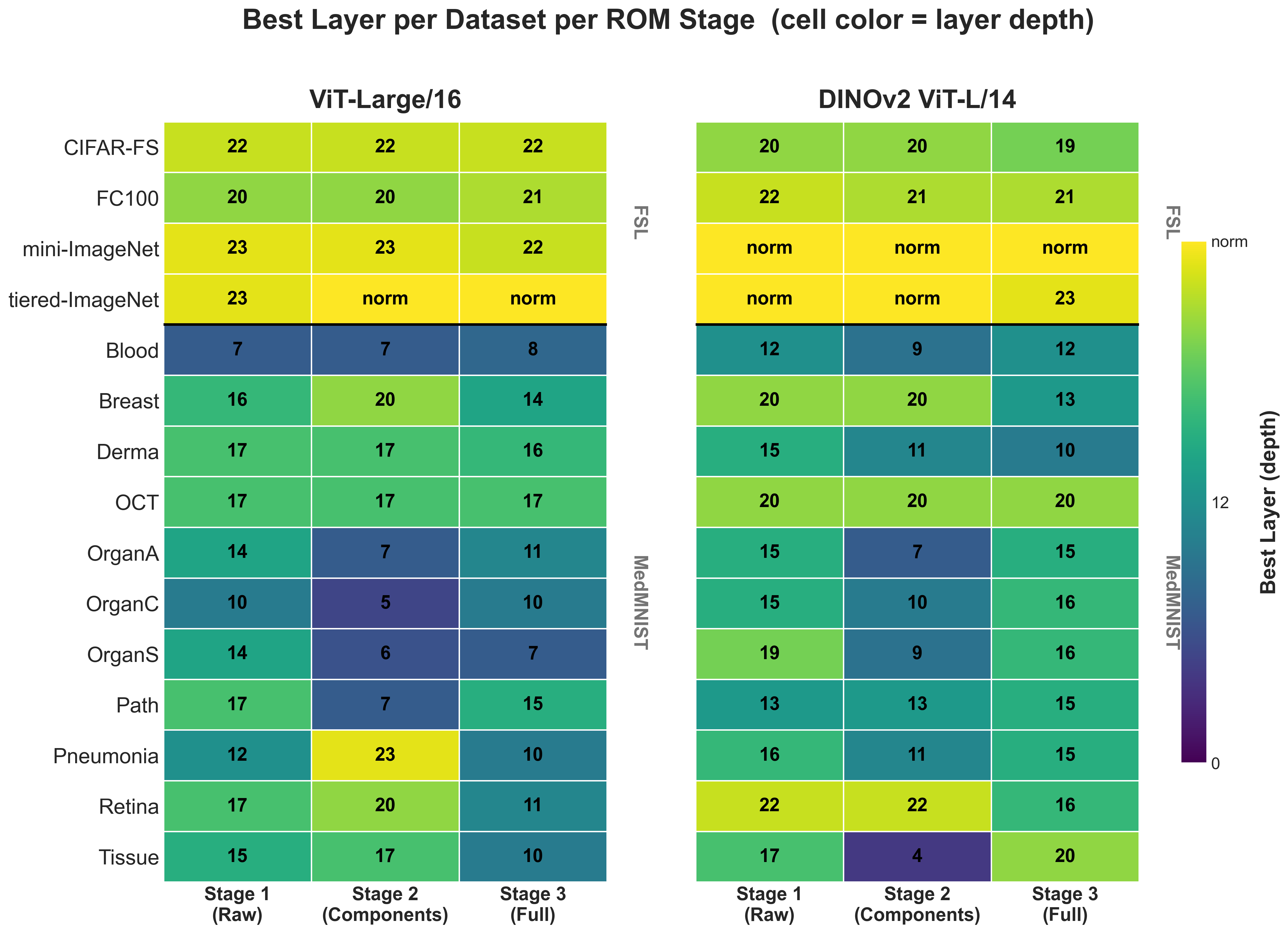}
    \caption{Heatmap visualization mapping the optimal feature extraction layer selected per dataset across the three operational encoding stages. The analysis evaluates the ViT-Large/16 (left) and the DINOv2 ViT-L/14 (right). Cell labels explicitly denote the specific transformer block index that yielded peak accuracy for that configuration. Cell shading corresponds to layer depth along a continuous gradient from the initial layers.}
    \label{fig:top_layer_heatmap}
\end{figure}

The layer-wise heatmap visualization in Figure \ref{fig:top_layer_heatmap} exposes a fundamental structural bifurcation in feature maturation across transformer architectures. For open-domain FSL benchmarks, the optimal representational depth is highly uniform, concentrating almost exclusively in the networks' final layers. Both ViT-Large/16 and DINOv2 ViT-L/14 consistently maximize downstream accuracy at deep penultimate layers or within the post-backbone normalization block. 

In sharp contrast, the optimal extraction layers for the MedMNIST v2 exhibit a highly dispersed, modality-dependent distribution spanning early, intermediate, and late layers. Specialized clinical tasks requiring structural geometry, such as \textit{Blood}, \textit{OrganA}, and \textit{OrganC}, achieve peak performance at early-to-middle layers, where granular primitives, textures, and edge boundaries remain unaggregated. Conversely, more abstract diagnostic tasks like \textit{Breast}, \textit{OCT}, and \textit{Retina} demand significantly deeper representations. This stark divergence demonstrates that a rigid, fixed-layer extraction policy is inherently sub-optimal for specialized domains; rather, the optimal Aristotelian representational depth is highly dynamic and intrinsically tied to the structural properties of each specific imaging modality.

\clearpage

\begin{figure}[htbp]
    \centering
    \includegraphics[width=0.9\linewidth]{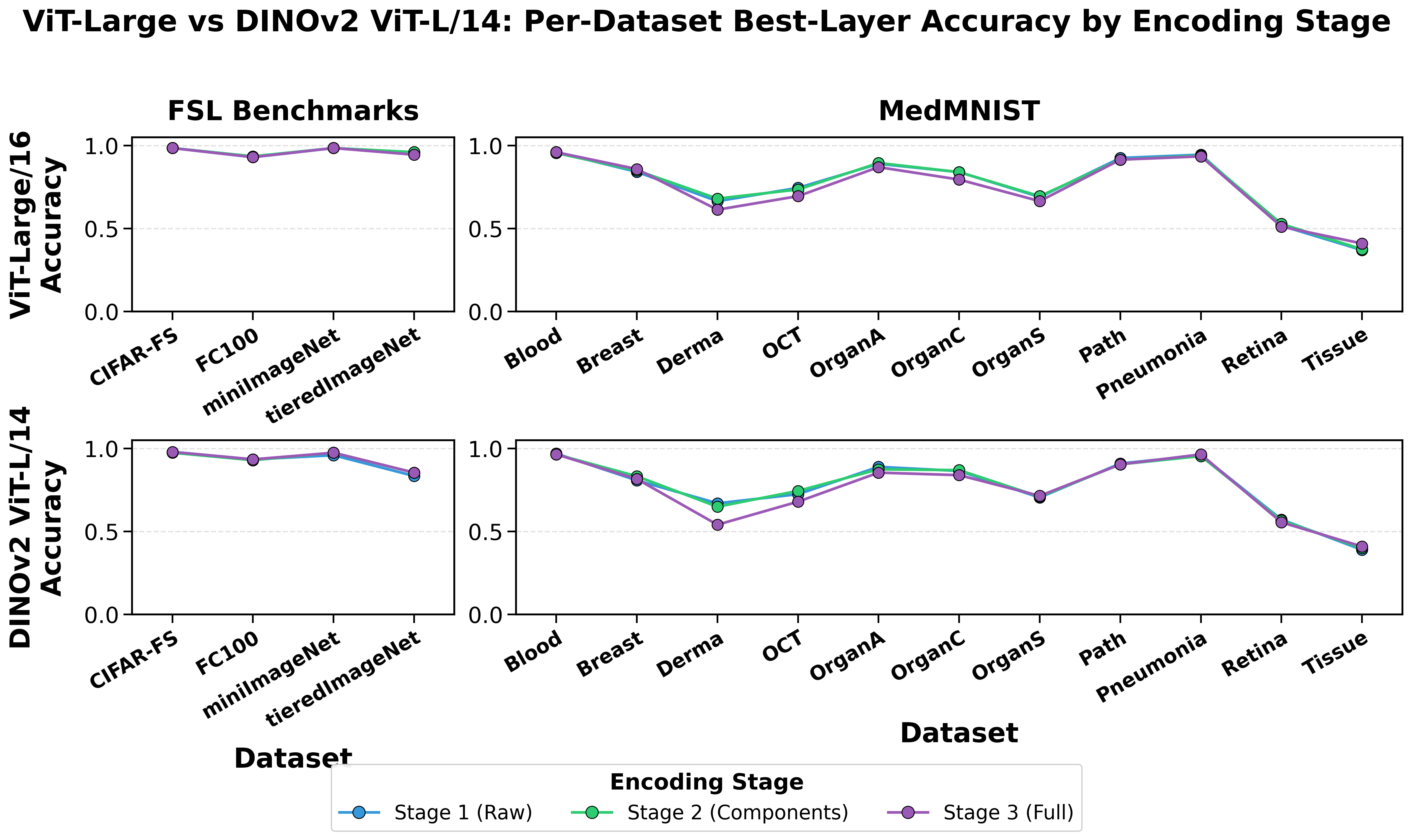}
    \caption{Per-dataset peak accuracies evaluated at the best-performing layer across the three encoding stages. The top row tracks a standard, supervised/self-supervised \text{ViT-Large/16} backbone, while the bottom row tracks the self-supervised \text{DINOv2 ViT-L/14} architecture. Evaluation is split side-by-side between the open-domain FSL Benchmarks (left) and the specialized MedMNIST v2 suite (right).}
    \label{fig:per_data_acc}
\end{figure}

The empirical evaluation presented in Figure \ref{fig:per_data_acc} maps the per-dataset peak accuracy achieved at each model's optimal layer across three  encoding stages. This analysis benchmarks the ViT-Large/16 (top row) and the DINOv2 ViT-L/14 (bottom row), evaluated across both FSL benchmarks and MedMNIST v2.

A striking empirical takeaway from \ref{fig:per_data_acc} is the extreme stability of downstream classification performance across all three processing stages. For the FSL benchmarks (left column), the accuracy trajectories for all stages remain virtually identical and heavily overlapping. This performance preservation is remarkably consistent across both the ViT-Large/16 and the DINOv2 architecture. It indicates that the primary discriminative manifolds are not disrupted by aggressive feature compression or vocabulary cluster quantization.

This topological resilience is equally evident across the structurally complex MedMNIST v2 suite (right column). Across highly disparate clinical imaging modalities the geometric alignment holds firm. Minor variations manifest exclusively in a small subset of modalities (such as \textit{Derma} and \textit{OCT}), where the localized linear compression of Stage 2 provides a marginal accuracy benefit over Stage 1 and Stage 3. 

Ultimately, the tight convergence of these performance profiles in \ref{fig:per_data_acc} validates a key structural property of Aristotelian Manifolds: the underlying semantic boundaries engineered by high-capacity foundation model backbones are inherently stable. The preservation of accuracy from Stage 1 through Stage 3 confirms that researchers can utilize aggressive dimensionality reduction without sacrificing classification accuracy.

\clearpage
\section{MedMNIST Few-Shot}

\begin{figure}[htbp]
    \centering
    \includegraphics[width=0.9\linewidth]{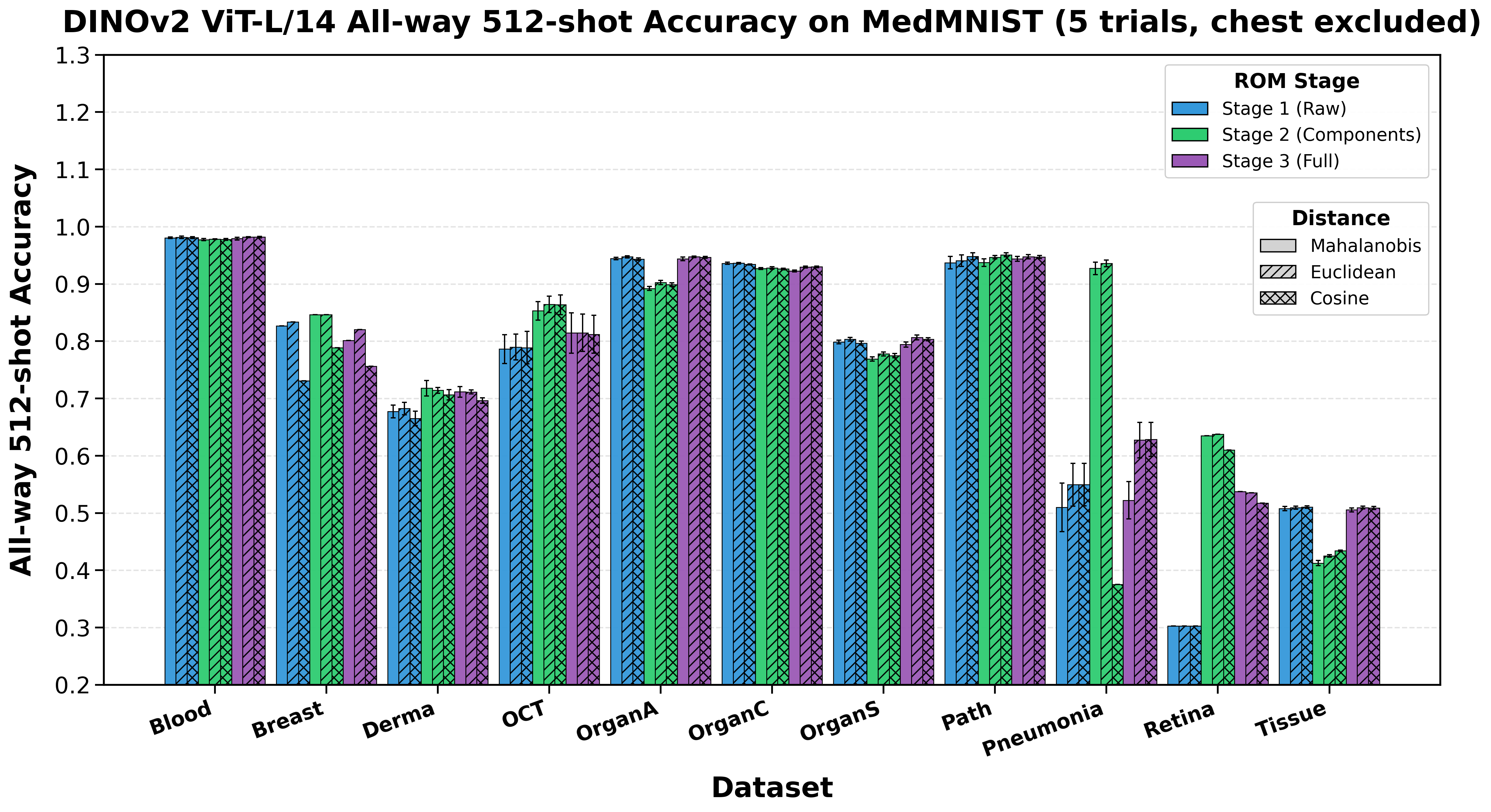}
    \caption{Mean all-way 512-shot classification accuracy across MedMNIST v2 (excluding multi-label ChestMNIST) evaluated using a frozen DINOv2 ViT-L/14 backbone. Performance is broken down across the three encoding stages and distance metrics. Bar heights show the mean computed across the 11 individual dataset accuracy scores across the 5-trials. Error bars denote the standard deviation ($\pm$ SD) across the 5-trials.}
    \label{fig:dinov2_medmnist_512shot}
\end{figure}

An essential dimension of characterizing Aristotelian Manifolds involves evaluating how their underlying geometric boundaries withstand different proximity constraints and aggressive representation mapping. Figure \ref{fig:dinov2_medmnist_512shot} profiles the mean all-way 512-shot classification accuracy across 11 multi-domain MedMNIST v2 datasets utilizing the DINOv2 ViT-L/14 backbone. Performance is cross-examined across the three encoding stages paired with three distinct distance metrics.

A macro-level inspection of the performance profiles reveals that for highly structured, saturated tasks, such as \textit{Blood}, \textit{OrganA}, and \textit{OrganC}, the choice of downstream distance metric and encoding stage yields almost negligible variance. 

However, highly volatile and distinct trends emerge within specialized clinical tasks characterized by more subtle topological anomalies, most notably in \textit{Pneumonia} and \textit{Retina}. In the \textit{Pneumonia} cohort, Stage 1 features exhibit catastrophic degradation when paired with Cosine distance. Crucially, moving to Stage 2 with a Mahalanobis or Euclidean metric completely corrects this drop. This behavior underscores a vital property of the Aristotelian framework: low-rank linear component decomposition acting on a targeted task subspace can isolate and amplify faint, critical diagnostic signals that are otherwise overwhelmed by unweighted feature matrices or global directional metrics like Cosine.

Ultimately, these empirical observations confirm that Aristotelian Manifold optimization cannot rely on a generic, one-size-fits-all metric policy. While standard Euclidean or Cosine distance metrics suffice for generic tasks, specialized medical modalities with high domain-shift variables require an explicit, intentional coupling of low-rank component extraction (Stage 2) and covariance-aligned distance metrics to reveal their optimal discriminative structure.

\clearpage
\section{Layer-wise}

This section presents the comprehensive layer-wise classification accuracy profiles across our entire evaluation suite. The subsequent figures track performance across four major architectural paradigms: VGG, ResNet, ViT, and DINOv2. 

For each architecture family, performance is cross-examined across both open-domain Few-Shot Learning (FSL) benchmarks and the specialized MedMNIST v2. Trajectories are plotted layer-by-layer across all three encoding stages to illustrate how downstream feature utility evolves as a function of network depth and representation compression.

\subsection{FSL Benchmarks}

  \begin{figure}[htbp]
    \centering
    \includegraphics[width=0.9\linewidth]{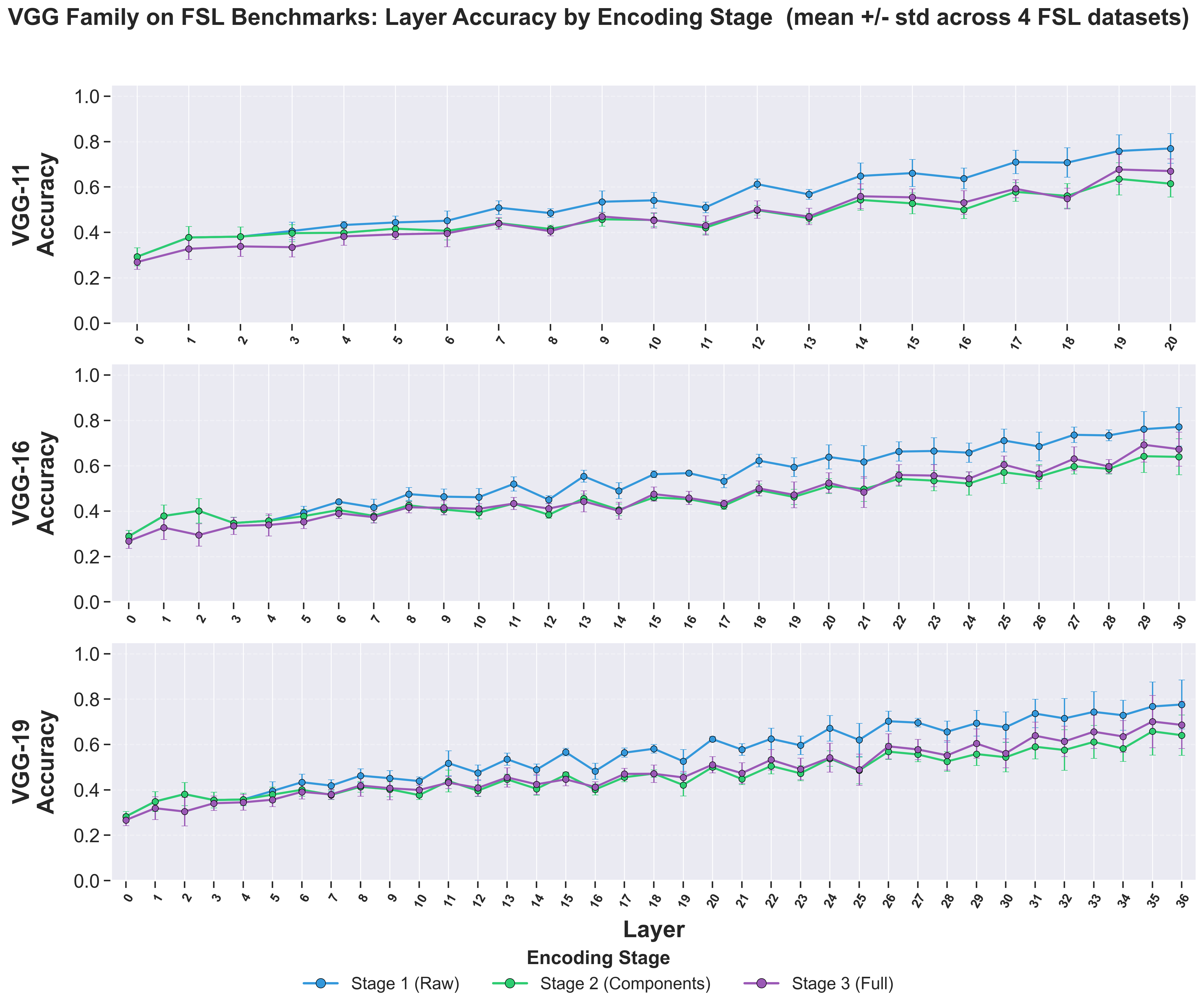}
    \caption{Layer-wise classification accuracy of VGG architectures on FSL benchmarks.}
    \label{fig:fsl_vgg_layers}
\end{figure}

  \begin{figure}[htbp]
    \centering
    \includegraphics[width=0.9\linewidth]{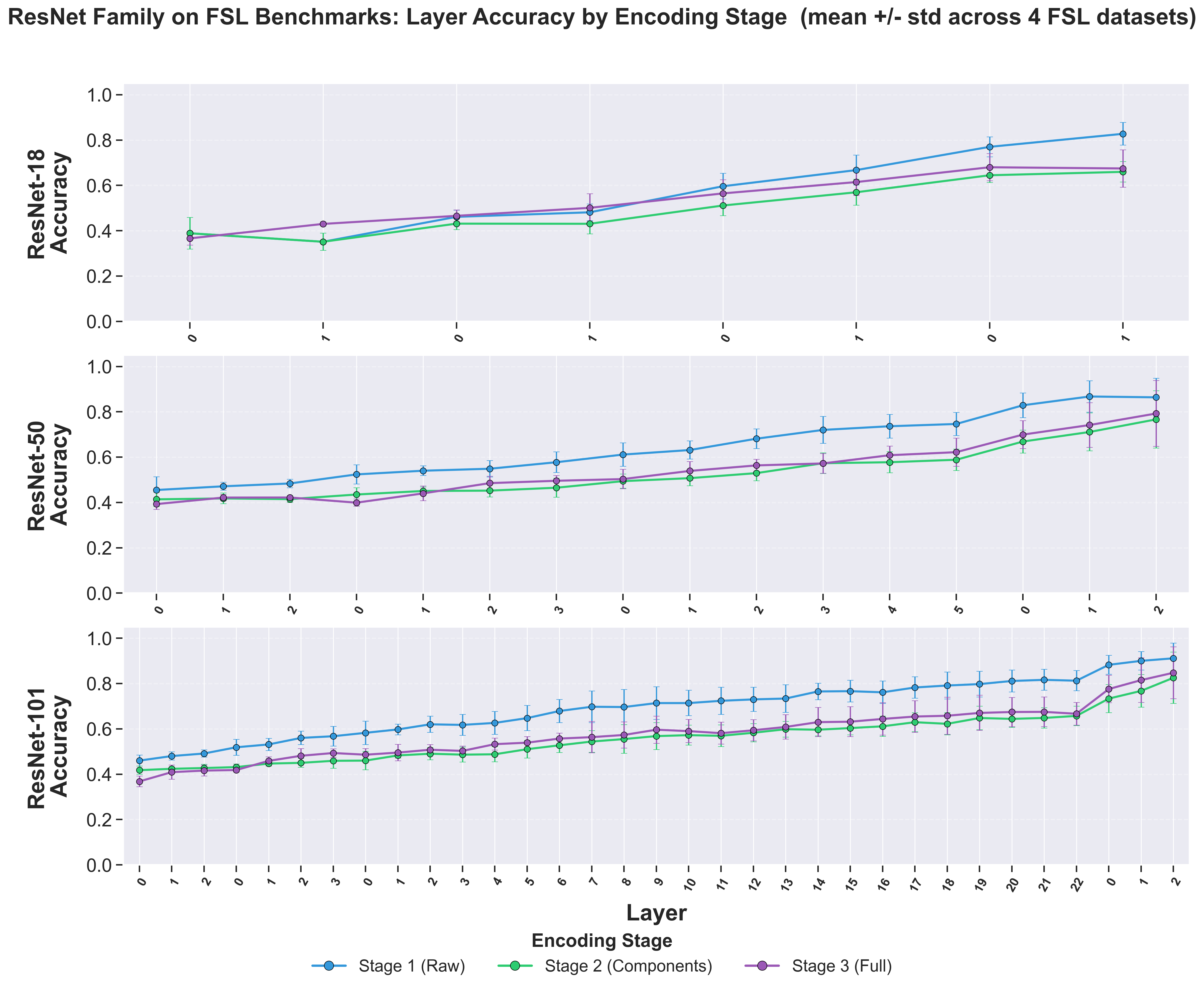}
    \caption{Layer-wise classification accuracy of ResNet architectures on FSL benchmarks.}
    \label{fig:fsl_resnet_layers}
\end{figure}

  \begin{figure}[htbp]
    \centering
    \includegraphics[width=0.9\linewidth]{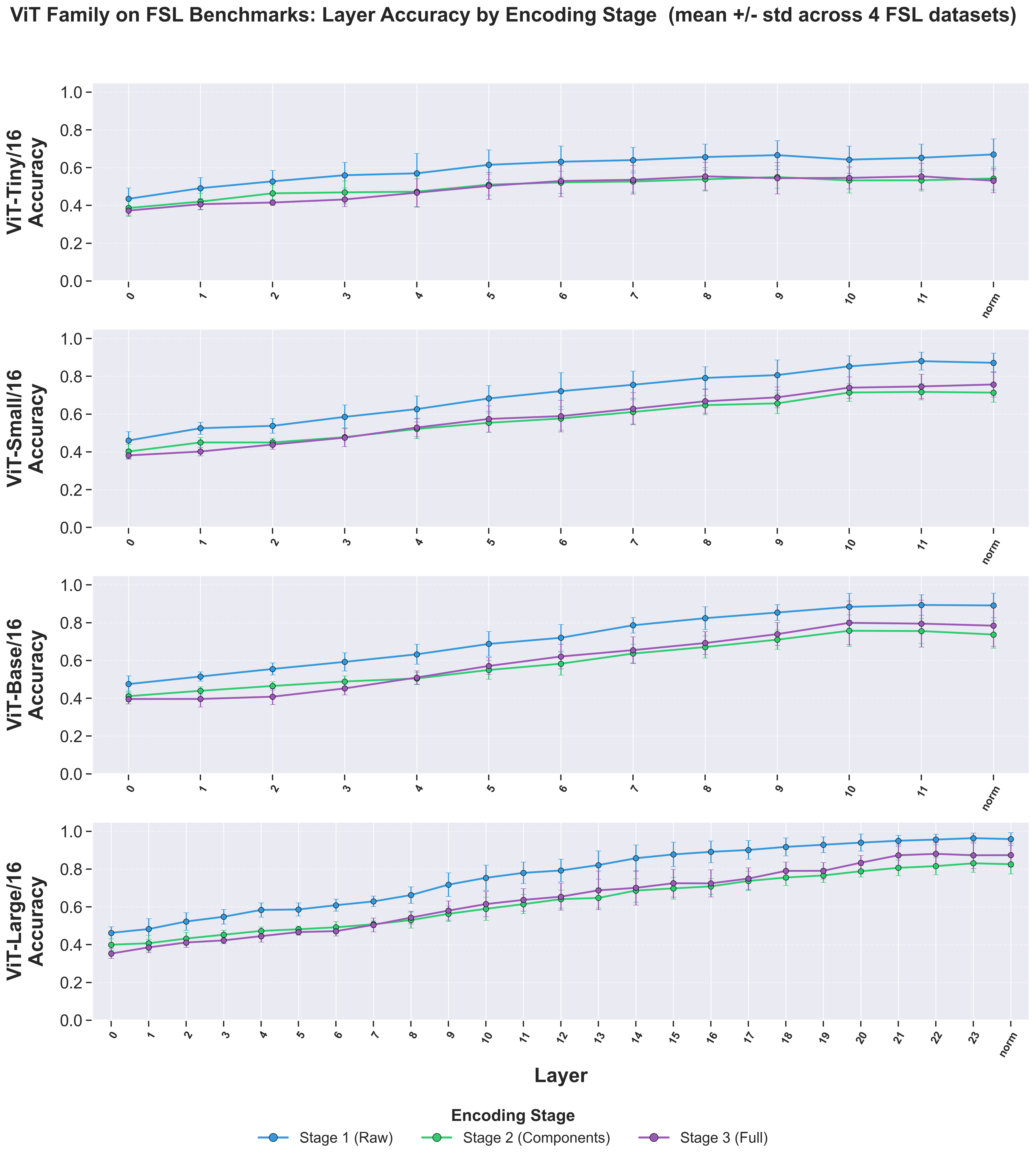}
    \caption{Layer-wise classification accuracy of ViT architectures on FSL benchmarks.}
    \label{fig:fsl_vit_layers}
\end{figure}

  \begin{figure}[htbp]
    \centering
    \includegraphics[width=0.9\linewidth]{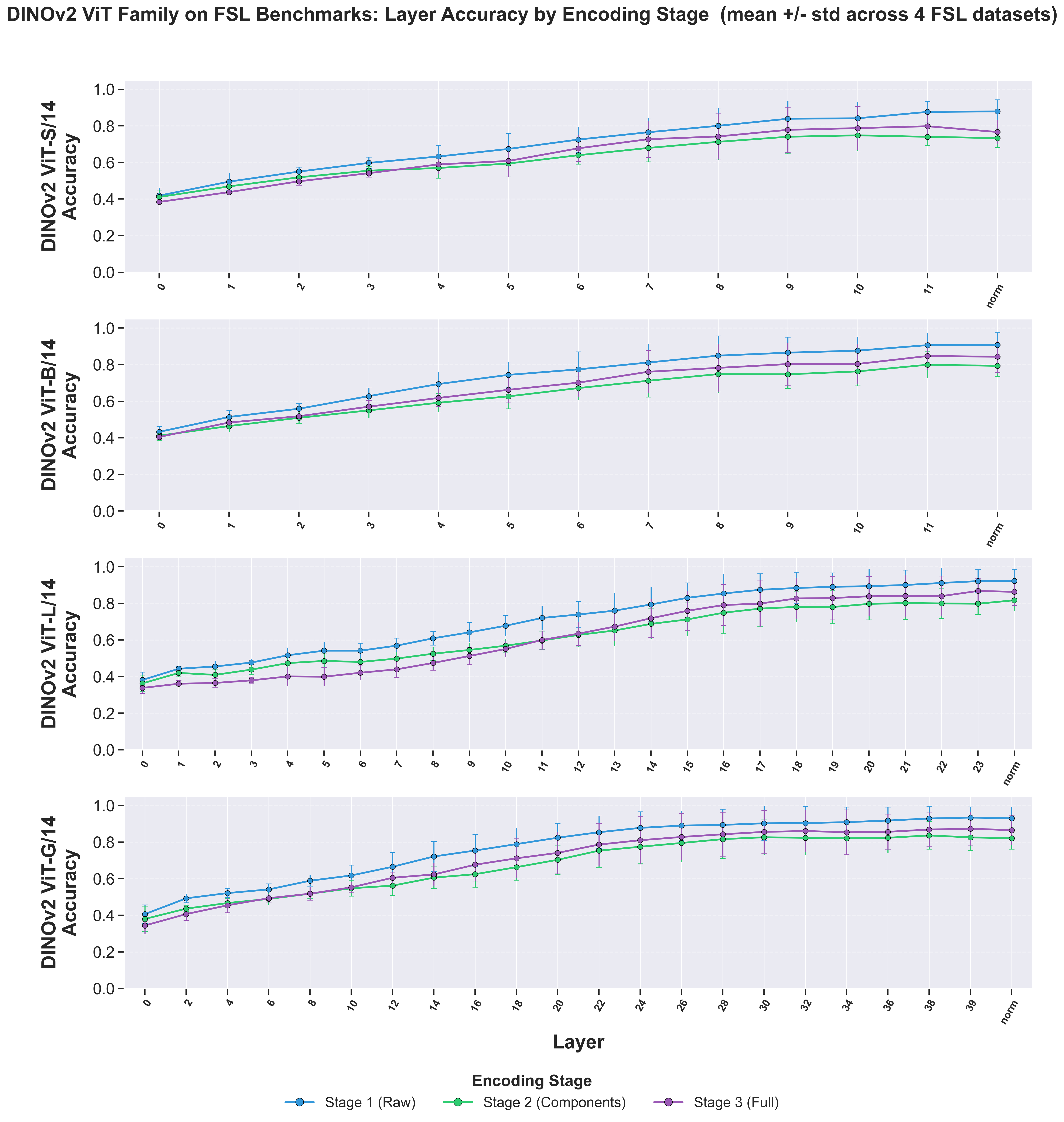}
    \caption{Layer-wise classification accuracy of DINOv2 architectures on FSL benchmarks.}
    \label{fig:fsl_dinov2_layers}
\end{figure}

\clearpage
\subsection{MedMNIST}
  \begin{figure}[htbp]
    \centering
    \includegraphics[width=0.9\linewidth]{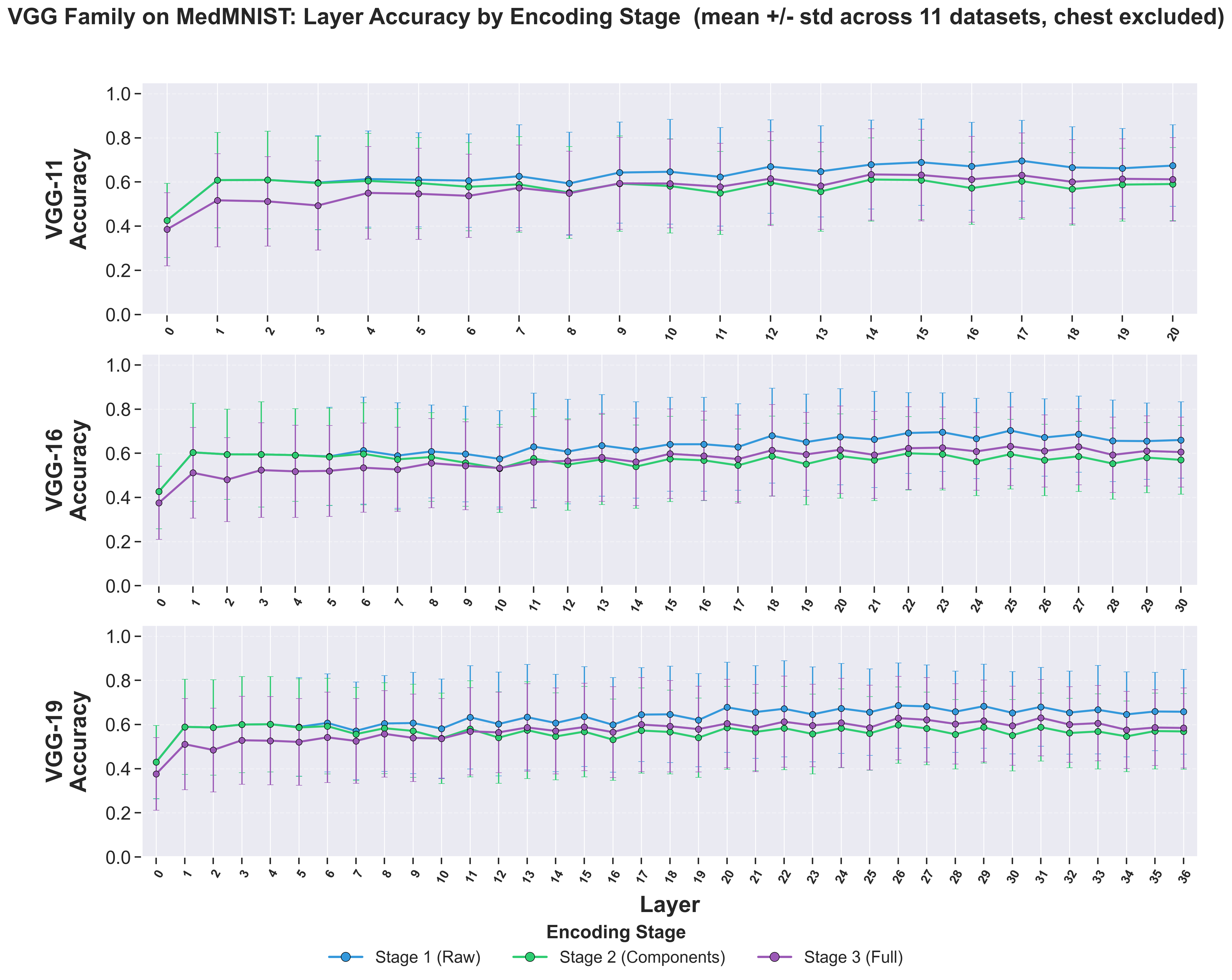}
    \caption{Layer-wise classification accuracy of VGG architectures on MedNIST v2.}
    \label{fig:med_vgg_layers}
\end{figure}

  \begin{figure}[htbp]
    \centering
    \includegraphics[width=0.9\linewidth]{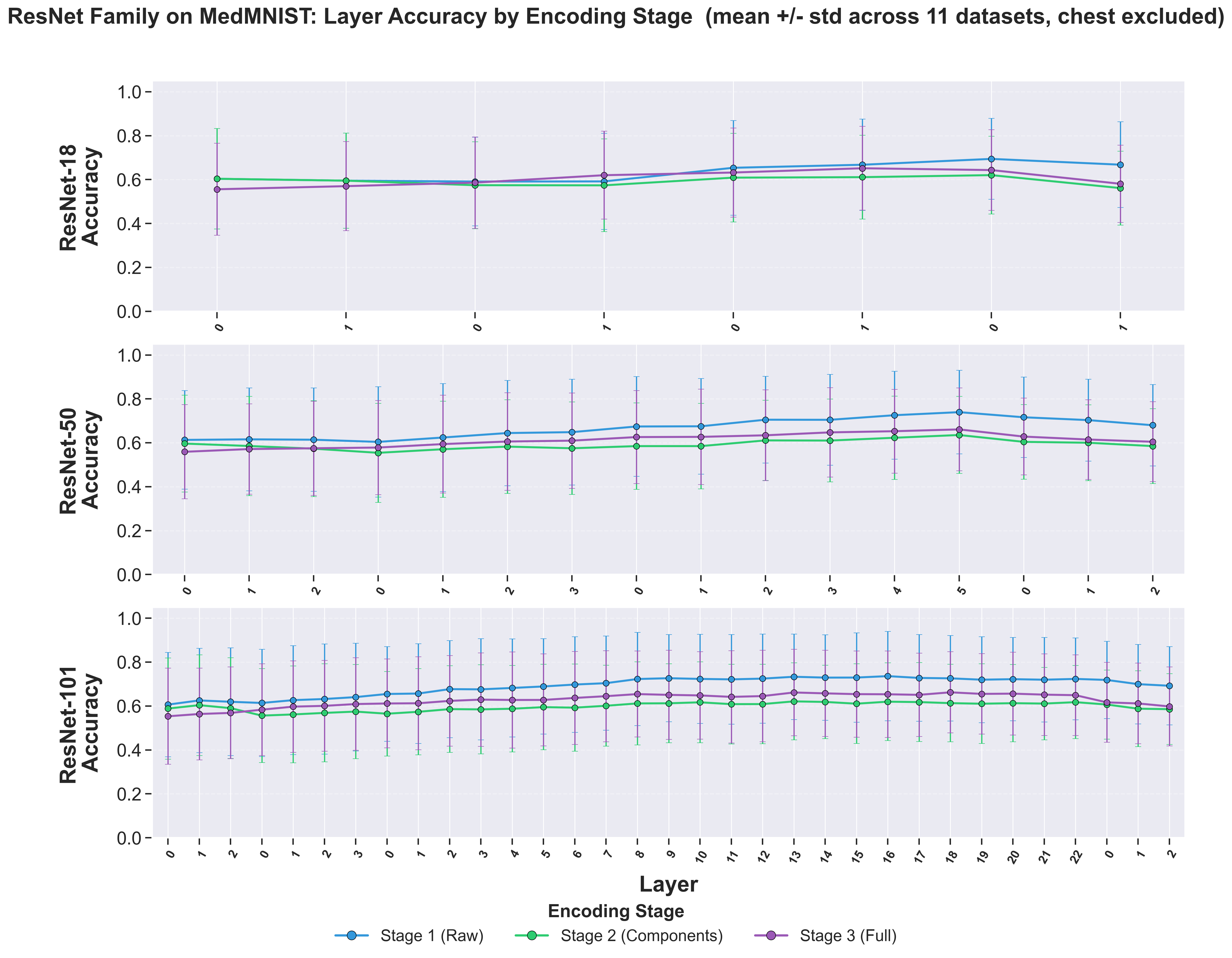}
    \caption{Layer-wise classification accuracy of ResNet architectures on MedNIST v2.}
    \label{fig:med_resnet_layers}
\end{figure}

  \begin{figure}[htbp]
    \centering
    \includegraphics[width=0.9\linewidth]{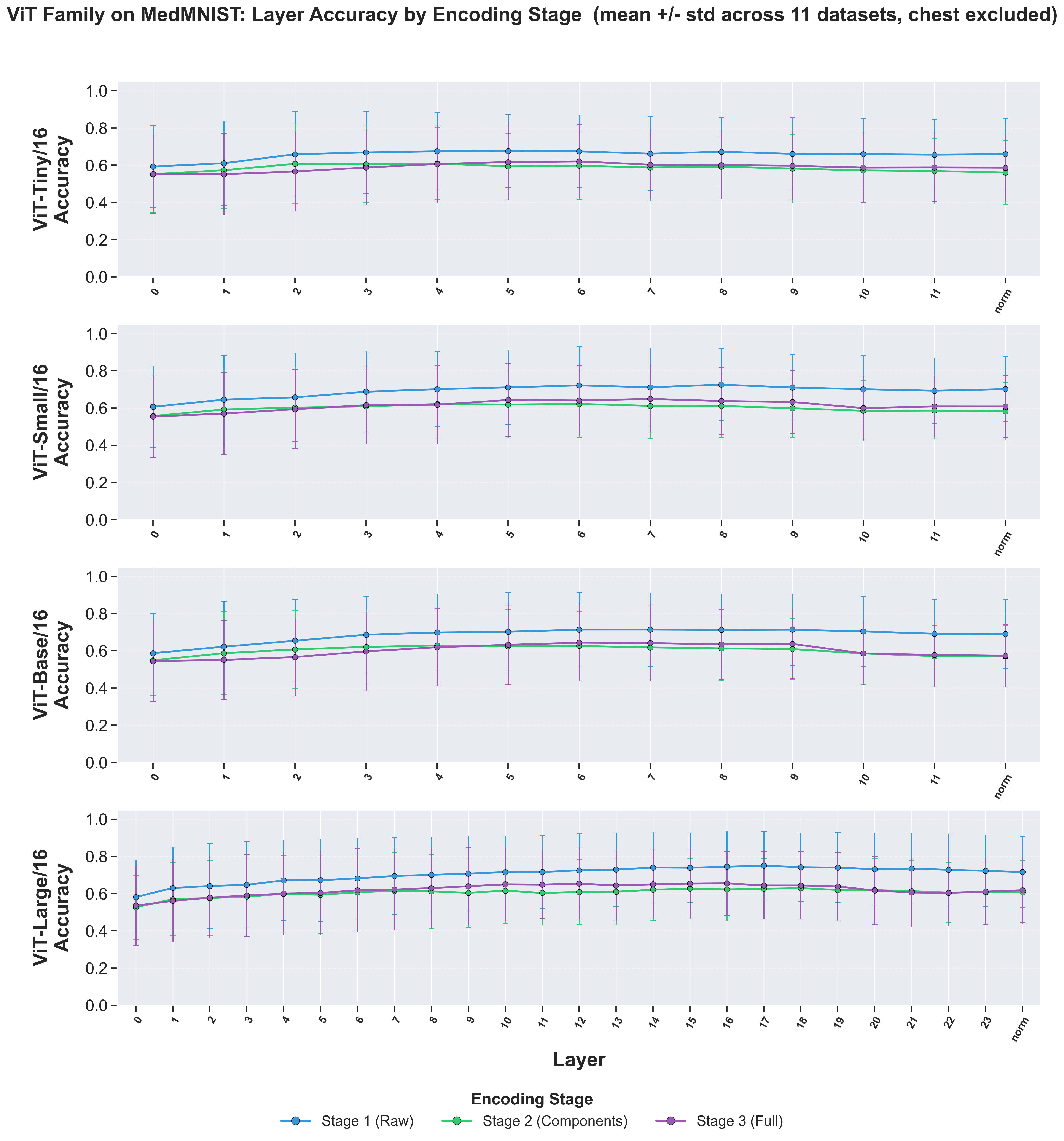}
    \caption{Layer-wise classification accuracy of ViT architectures on MedNIST v2.}
    \label{fig:med_vit_layers}
\end{figure}

  \begin{figure}[htbp]
    \centering
    \includegraphics[width=0.9\linewidth]{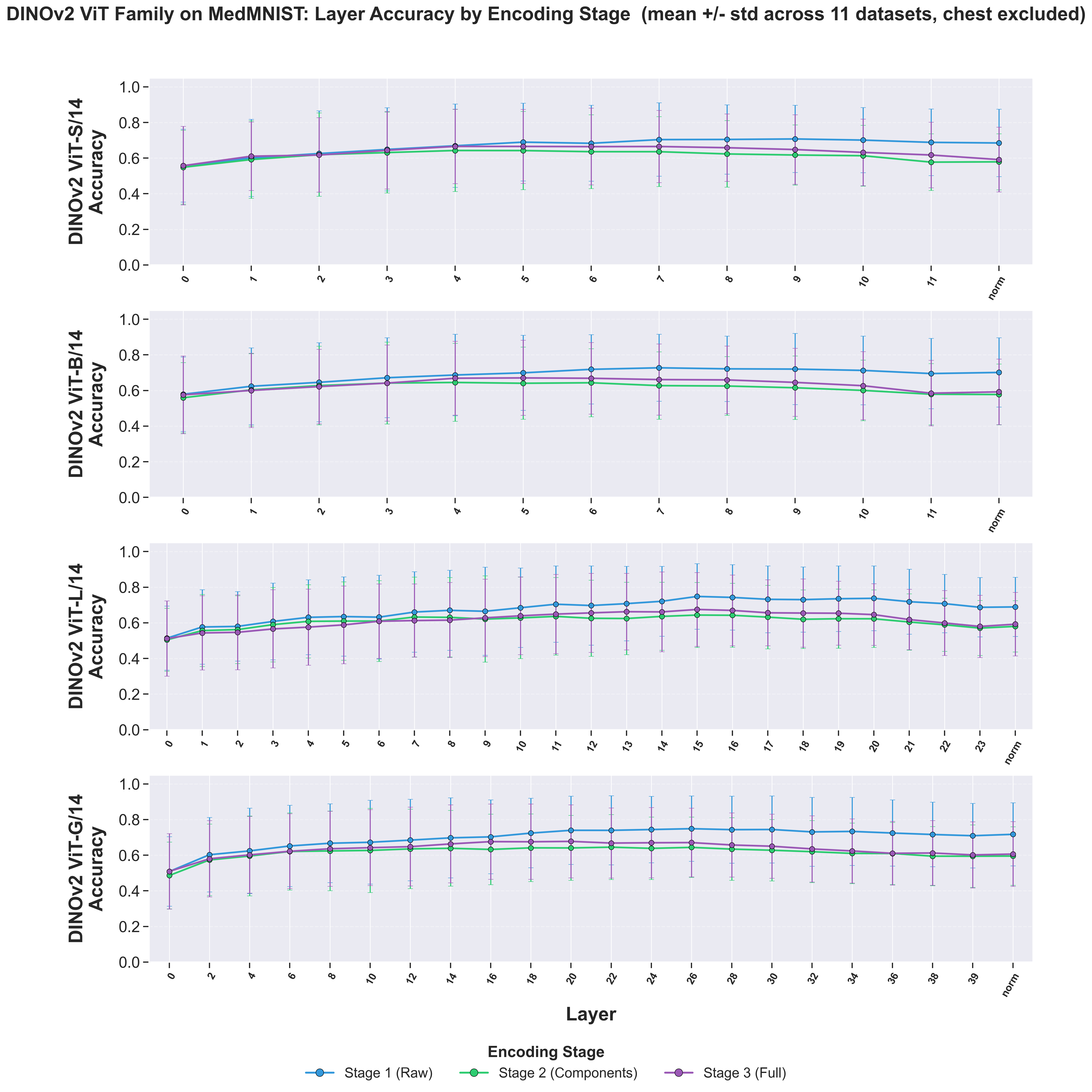}
    \caption{Layer-wise classification accuracy of DINOv2 architectures on MedNIST v2.}
    \label{fig:med_dino_layers}
\end{figure}

\end{document}